\documentclass[10pt]{article}

\usepackage[utf8]{inputenc}
\usepackage[T1]{fontenc}
\usepackage{amsmath,amssymb,amsfonts}
\usepackage{graphicx}
\graphicspath{{figures/}{./}}
\usepackage{booktabs}
\usepackage{array}
\usepackage{multirow}
\usepackage{hyperref}
\usepackage{xcolor}
\usepackage{listings}
\usepackage{caption}
\usepackage{subcaption}
\usepackage{algorithm}
\usepackage{algorithmic}
\usepackage{balance}
\usepackage{float}
\usepackage{geometry}
\usepackage{url}

\newcommand{\mycode}[1]{\texttt{\detokenize{#1}}}

\lstdefinestyle{mermaid}{
  basicstyle=\ttfamily\footnotesize,
  breaklines=true,
  frame=single,
  backgroundcolor=\color{gray!8},
  keywordstyle=\bfseries,
  showstringspaces=false,
  numbers=none,
  escapeinside={(*@}{@*)}
}

\title{%
  \includegraphics[height=1cm]{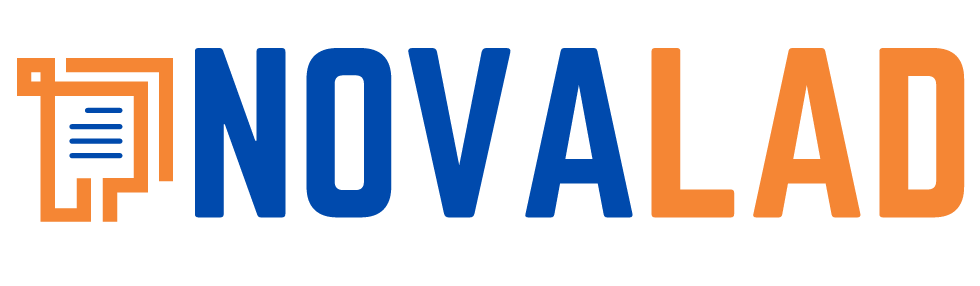}\\[1em]
  \textbf{NovaLAD: A Fast, CPU-Optimized Document Extraction Pipeline for Generative AI and Data Intelligence}}
\author{Aman Ulla\\\texttt{connectamanulla@gmail.com}}
\date{}

\begin{document}
\maketitle

\begin{abstract}
Document extraction is an important step before retrieval-augmented generation (RAG), knowledge bases, and downstream generative AI can work. It turns unstructured documents like PDFs and scans into structured text and layout-aware representations. We introduce NovaLAD, a comprehensive document parsing system that integrates two concurrent YOLO object detection models—element detection and layout detection—with rule-based grouping and optional vision-language enhancement. When a page image is sent in, the first thing that happens is that it goes through both models at the same time. The element model finds semantic content like the title, header, text, table, image, and so on, and the layout model finds structural regions like layout\_box, column\_group, multi\_column, row\_group, and so on. A key design decision is to first send an image or figure through an image classifier (ViT) that decides whether it is relevant or not. Only useful images are then submitted to the Vision LLM for title, summary, and structured information, which cuts down on noise and costs. NovaLAD is built for speed: it works on CPU, employs parallel execution for detection, classification, OCR, and conversion, and generates several forms, including structured JSON, Markdown, RAG-ready texts, and knowledge graphs. We test on the DP-Bench benchmark (upstage/dp-bench) and get \textbf{96.49\% TEDS} and \textbf{98.51\% NID}, which is better than both commercial and open-source parsers. This paper explains how to extract data, how the architecture works, how data flows, and how to make NovaLAD both accurate and usable without needing a GPU.
\end{abstract}

\section{Introduction}

\subsection{The Role of Extraction in AI and Data Intelligence}

Document extraction, also known as document parsing, is the process of turning complicated documents like PDFs, scanned images, and office files into structured, machine-readable formats like JSON, HTML, or Markdown while keeping the layout, reading order, and meaning of each element \cite{dpbench}. It sits between all of the cognitive systems that come after basic enterprise content.

Generative AI and RAG. Retrieval-augmented generation (RAG) systems depend on chunked and labeled document material for retrieval. If the extraction is bad, the chunks will be out of order, the structure (headings, tables, lists) will be lost, and the text will be duplicated or absent. This will immediately lower the quality of the answers and increase the number of hallucinations \cite{rag}. So, the quality of the extraction is a key factor in how well RAG works.

Data intelligence and analysis. Dashboards, NLP models, and knowledge graphs get their information from tables, figures, and metadata taken from reports and filings. Extraction that is not consistent or full requires expensive manual adjustment and makes automation less effective. To compare parsers and move the field forward, we need standardized benchmarks and metrics, like NID for layout and TEDS for tables.

Why it matters. Extraction is not just a step before processing; it is a \emph{core capability} for AI and data intelligence. It sets the quality of the signal that all the models that come after it get. Enhancing extraction directly benefits all applications developed upon it. In businesses, documents often come as PDFs or scans that don't have a built-in structure. Without strong extraction, RAG systems pull out irrelevant or broken-up parts, and analytics pipelines need to be manually normalized. NovaLAD solves this problem by giving you a single pipeline that keeps the reading order, layout semantics, and multi-format output for both benchmarking and production use.

\subsection{Contributions and Paper Outline}

We describe \textbf{NovaLAD}, a document extraction pipeline that:
\begin{itemize}
  \item Does "parallel dual YOLO detection": when a page image is input, it is first sent to both a "element model" that finds the title, header, page header, text, table, image, etc., and a "layout model" that finds the layout box, column group, column text, group, multi-column, and row group. It also does OCR with EasyOCR. For detected "image/figure" elements, it uses a "image classifier" (ViT) to label them as useful or not, and only sends useful images (and tables) to the Vision LLM (e.g., GPT-4o-mini) for title, summary, and structured content.
  \item Makes layout-aware grouping (multi-column, row, and reading-order groups) and serialization that works for benchmarks and RAG.
  \item It can output structured JSON (DP-Bench compatible), Markdown, LangChain-style documents, and knowledge graphs all at once.
  \item Is optimized for \textbf{speed} and \textbf{CPU execution}: parallel threads for detection, image classification, and conversion; no mandatory GPU.
\end{itemize}

We evaluate on the public \textbf{DP-Bench} dataset (upstage/dp-bench on Hugging Face), reporting TEDS, TEDS-S, and NID. NovaLAD achieves state-of-the-art \textbf{TEDS 96.49\%} and \textbf{NID 98.51\%} on this benchmark.

The rest of the paper is organized as follows: Section~\ref{sec:related} briefly positions our work; Section~\ref{sec:method} details the NovaLAD architecture and extraction method; Section~\ref{sec:eval} describes DP-Bench and results; Section~\ref{sec:optimization} discusses optimization and CPU-based design; Section~\ref{sec:conclusion} concludes.

\section{Background and Related Work}
\label{sec:related}

\paragraph{Rule-based and classical approaches.}
Early document parsing used the structure of PDFs (including embedded fonts and content streams) and geometric heuristics to figure out blocks, columns, and the order in which to read them \cite{constraint_layout, pdfminer}. PDFMiner and Apache PDFBox are two examples of tools that can get text and locations, although they don't always work on scanned documents or documents with complicated layouts. The recovery of reading order has been conceptualized as a constraint-satisfaction or ordering problem concerning identified regions \cite{reading_order_survey}.

\paragraph{Learning-based layout analysis.}
Deep learning has made it easier to find and sort layouts. CNN-based algorithms can guess bounding boxes and labels for headings, paragraphs, figures, and tables \cite{layoutlm, docbank}. LayoutLM \cite{layoutlm} and its derivatives integrate text and 2D layout into a singular transformer for document comprehension. Faster R-CNN, DETR, and YOLO are all object detection frameworks that have been adapted to work with document element detection \cite{yolo_doc, detr_doc}. NovaLAD employs YOLOv10 to find both elements and layouts. This means that each model only needs to do one forward pass, which makes CPU inference faster.

\paragraph{Table structure recognition.}
Tables are among the most challenging elements. Tree Edit Distance based metrics (TEDS, TEDS-S) \cite{teds} measure structural and content similarity. Methods range from graph neural networks and transformers to dedicated table structure recognition models \cite{table_transformer, tablestr}. NovaLAD combines detection of table regions with EasyOCR for cell content and optional vision-language models for semantic summaries, then outputs structure-preserving JSON and Markdown suitable for TEDS evaluation.

\paragraph{Commercial and open-source parsers.}
Commercial APIs like AWS Textract \cite{textract}, Google Document AI \cite{docai}, and Microsoft Document Intelligence \cite{ms_read} are quite accurate, but they charge per page and take a long time to process. They could also need to be deployed on a GPU or in the cloud. Unstructured is one of the open-source options.io (chunking based on heuristics and models), Llamaparse (parsing with LLM), and several packages for converting PDFs to text \cite{unstructured}. DP-Bench \cite{dpbench} offers a standardized benchmark (NID, TEDS, TEDS-S) on a static dataset, facilitating direct comparison. NovaLAD is a single, open pipeline that beats these systems on DP-Bench while still being optimized for CPUs and able to work with several formats.

\paragraph{Reading order and grouping.}
For NID and RAG chunk quality, it is very important to serialize items in the right order. Multi-column and table layouts need clear criteria for grouping and ordering, or learned order predictors \cite{reading_order_survey}. NovaLAD combines layout-model regions (such multi-column, row, and group) along with DBSCAN and geometric sorting to put items into groups. It then merges groups with non-group elements based on their vertical location, creating a stable reading-order sequence without needing another learnt module.

\section{NovaLAD Pipeline}
\label{sec:method}

\subsection{Pipeline Overview}

\begin{figure*}[t]
\centering
\includegraphics[width=500pt]{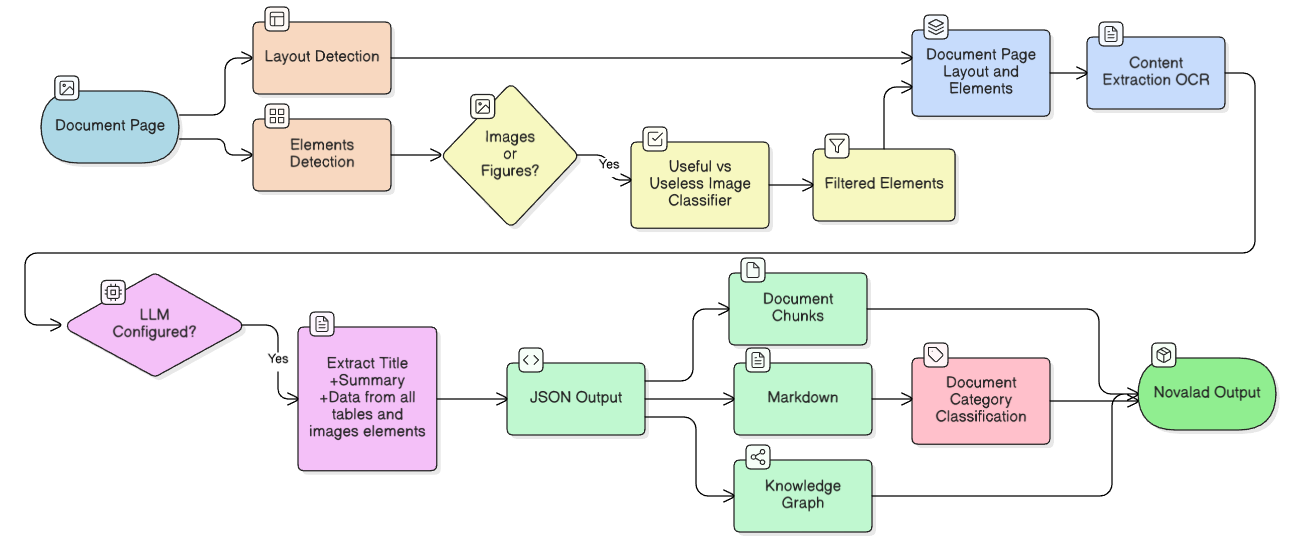}
\caption{Complete end-to-end NovaLAD pipeline. The process begins with a document page fed to parallel Layout Detection and Elements Detection (YOLO models). Images/figures are filtered by a useful vs useless classifier. Layout and filtered elements merge into a unified representation, followed by OCR content extraction. If an LLM is configured, advanced extraction of titles, summaries, and data is performed. The pipeline outputs structured JSON, which is converted to multiple formats (document chunks, Markdown, knowledge graph) with document category classification, culminating in the final NovaLAD output.}
\label{fig:process-flow}
\end{figure*}

The NovaLAD extraction process changes the pages of a document into organized, multi-format outputs through a sequence of steps that are all connected. Figure~\ref{fig:process-flow} shows the whole pipeline from start to finish. The first step is to feed a \textbf{Document Page} to two simultaneous YOLO object identification models: \textbf{Layout identification} and \textbf{Elements Detection}. An Image Classifier (ViT) sorts detected elements into useful and useless images, creating Filtered Elements. The layout data and the filtered elements are combined to make a single representation of the Document Page Layout and Elements. Content Extraction OCR gets text from all areas. If a LLM is set up, images and tables can have extracted titles, summaries, and structured data added to them. The pipeline makes a JSON Output, which is then changed at the same time into three different formats: Document Chunks (for RAG), Markdown, and Knowledge Graph. Finally, the document is sorted into a category by Document Category Classification, and all the outputs come together to make the final NovaLAD Output. The next subsections go into further depth about each step.

\subsection{Stage 1: Parallel Object Detection}

The first step in the NovaLAD pipeline uses two YOLO-based object identification models that operate at the same time: Layout identification and Element Detection. Both models work on the same rendered page image (300 DPI) at the same time using ThreadPoolExecutor, which makes it possible to do inference quickly at the same time. The outputs from each models are separate and are combined subsequently.

\subsubsection{Layout Detection}

The \textbf{Layout Detection model} (YOLOv10) identifies structural regions that define the spatial organization of the page. It predicts bounding boxes for layout-level regions such as:

The \textbf{Layout Detection model} (YOLOv10) identifies structural regions that define the spatial organization of the page. It predicts bounding boxes for layout-level regions such as:

\begin{itemize}
  \item \texttt{layout\_box}: generic layout container
  \item \texttt{column\_group}: group of columns
  \item \texttt{column\_text}: single text column
  \item \texttt{group}: generic element grouping
  \item \texttt{multi\_column}: multi-column layout regions
  \item \texttt{row\_group}: horizontal row arrangement
\end{itemize}

The layout model was trained on a bespoke dataset for 30 epochs. The default value for the confidence threshold is 0.20. These areas give the structure for putting together the identified pieces later on (for example, multi-column text flows and row-based categories). Section~\ref{sec:eval-layout} shows the full training metrics and results.

\subsubsection{Element Detection}

The \textbf{Element Detection model} (YOLOv10) operates in parallel with layout detection and identifies semantic document elements---the actual content regions. It predicts bounding boxes and class labels for:

\begin{itemize}
  \item \texttt{title}: document or section titles
  \item \texttt{header}, \texttt{section}: section headings
  \item \texttt{page\_header}, \texttt{page\_footer}: repeating header/footer text
  \item \texttt{text}: body paragraph blocks
  \item \texttt{list\_item}: list items and bullets
  \item \texttt{table\_of\_content}: table of contents entries
  \item \texttt{table}: tabular data regions
  \item \texttt{image}: figures, charts, and images
  \item \texttt{table\_caption}, \texttt{image\_caption}: captions for tables and figures
\end{itemize}

The element model was trained for 50 epochs and did a great job of detecting all types of elements. The default value for the confidence threshold is 0.30. These discovered elements make up the semantic content that will be filtered, improved, and sent out. Section~\ref{sec:eval-elements} has all the information about the training results.

\subsection{Stage 2: Image Classification and Filtering}

After finding elements, all identified \texttt{image} and \texttt{figure} elements go through a \textbf{Image Classifier} that sorts them into two groups: \emph{Useful} and \emph{Useless}. Images that are useful are ones that have useful information in them, including charts, flowcharts, infographics, data visualizations, or other content that is worth transmitting to the Vision LLM. Decorative graphics, logos, ornamental features, and placeholder images that don't contribute any value are examples of "useless" images. This filtering step is very important for two reasons: (1) it keeps worthless photographs from getting in the way of the retrieved content, and (2) it cuts costs by only allowing useful images to be sent to Vision LLM.

\subsubsection{ViT Model Architecture}

The best-performing classifier in the end is a fine-tuned \textbf{Vision Transformer (ViT)} based on \mycode{google/vit-base-patch16-224-in21k}. The model architecture is ViTForImageClassification, and it has the following settings: image size $224 \times 224$, patch size 16, 12 transformer layers, hidden size 768, 12 attention heads, intermediate size 3072, and GELU activation. The model does single-label classification with two output classes: ``Useful'' (label id 0) and ``Useless'' (label id 1). We change the size of the input images to $224 \times 224$ and process them as sequences of $16 \times 16$ patches. Section~\ref{sec:eval-image-classifier} shows the outcomes of the training and testing.

Using \texttt{ThreadPoolExecutor}, the classifier runs at the same time on all the picture elements on a page. Images that are marked as "Useless" are kept in a list called "skipped\_images" and are not used in any downstream processing or output formats. Images that are marked as "Useful" (as well as all table elements, which skip this filter) go on to content extraction and optional LLM enrichment.

\subsection{Stage 3: Page Layout + Elements Integration}

The outputs from the parallel Layout Detection and Element Detection models, as well as the filtered elements from the image classifier, are combined to make a single \textbf{Page Layout + Elements} representation. This step puts each detected element in the right layout region, if there is one, depending on how they fit together and how they overlap.

The pipeline looks for all element boxes whose midpoints are inside the layout box for each layout bounding box (multi-column, row, or group). These elements make up a group that is linked to that layout area. Elements that don't fit into any layout region are considered as separate blocks. This spatial mapping makes a hierarchy: layout regions hold ordered lists of elements, which lets you serialize the reading order in multi-column and complex layouts.

\subsection{Stage 4: Content Extraction (OCR)}

When elements and layout regions are combined, OCR's "Content Extraction" feature pulls text from all identified areas. The pipeline takes a hybrid approach:

\begin{itemize}
  \item \textbf{PDF text layer} (via PyMuPDF): For text-like elements (title, section, text, list\_item, headers, footers), text is extracted directly from the PDF's embedded text layer using bounding-box queries (\texttt{textpage.get\_text\_bounded()}). This preserves formatting and avoids OCR errors when native text is available.
  \item \textbf{EasyOCR}: For table and image elements (and when PDF text is unavailable), the bounding box region is cropped from the rendered page image and passed to EasyOCR (English). The OCR result is stored as the element's text content.
\end{itemize}

Text normalization is done: titles and sections are cleaned (non-UTF-8 characters and too many punctuation marks are removed), and the body text is NFKC normalized, bullet points are standardized, and extra spaces are removed. An \textbf{Entity} represents each element. It has an id (UUID), a type (label), a confidence level, a value (\texttt{\{text: ...\}}), pixel coordinates (left, top, right, bottom), and computed properties (mid-point, x\_center, y\_center, and weight for ordering).

\subsection{Stage 5: LLM-Based Enrichment (Optional)}

If an LLM is set up (like GPT-4o-mini or GPT-4-vision), the pipeline can get deep content from photos and tables. For every "useful image" (those that passed the ViT classifier) and every table, the cropped base64 image of the element is given to a Vision LLM with a structured prompt (from insights/image.yaml) asking for:

\begin{itemize}
  \item \texttt{title}: a descriptive title for the image or table
  \item \texttt{summary}: a textual summary of the content
  \item \texttt{text} or \texttt{data}: for tables, structured data as a list of dictionaries (rows); for images, descriptive text or data extracted from charts/graphs
\end{itemize}

To reduce latency, LLM calls are done on each page at the same time. The JSON that was sent back is parsed and added to the entity's \texttt{value} field. This stage adds meaning to the data beyond what OCR can do (for example, by analyzing chart trends, extracting table structure, and characterizing the content of figures). It is especially useful for RAG systems that need to know what the visual information means in context.

If no LLM is configured, this stage is skipped, and only OCR-extracted text is retained.

\subsection{Stage 6: JSON Output and Multi-Format Export}

The extraction result is serialized to a \textbf{JSON Output} structure containing: 
\texttt{filename}, \texttt{total\_pages}, \texttt{total\-\allowbreak processed\-\allowbreak pages}, 
\texttt{metadata}, \texttt{document\_category}, and \texttt{pages} array. Each page includes \texttt{page\_number}, \texttt{elements} (dict of id $\to$ entity, in reading order), \texttt{groups} (layout regions with element IDs), \texttt{non\_groups} (elements not in any layout region), and \texttt{skipped\_images} (useless images excluded from output).

From this JSON, \textbf{four output formats} are generated \textbf{in parallel} via \texttt{ThreadPoolExecutor}:

\begin{enumerate}
  \item \textbf{Document Chunks (for RAG)}: Three types of chunks are produced: (a) page-level documents (all text concatenated per page), (b) header-based blocks (section/title plus following text/list items), and (c) per-element documents. Each chunk has \texttt{page\_content} and metadata (page number, element type, token count, filename, document category). Duplicates are removed by content hash. These chunks are ready for embedding and vector store ingestion in RAG pipelines.
  
  \item \textbf{Markdown}: Elements are serialized per page in reading order. Titles become \texttt{\#\#}, sections \texttt{\#\#\#}, text as paragraphs, lists as bullets (with extracted list items), and tables/images as Markdown tables (via pandas \texttt{to\_markdown}) or image links with captions/summaries. Headers/footers can be optionally skipped.
  
  \item \textbf{Knowledge Graph}: A graph is constructed with a root node (filename), page nodes, and one node per element (in reading order). Edges: root $\to$ page (\texttt{contains}), page $\to$ first element (\texttt{contains}), and between consecutive elements (\texttt{sibling} if same weight, \texttt{parent-child} if different weight based on schema). This graph captures document structure and semantic relationships for downstream knowledge base integration.
  
  \item \textbf{Document Category Classification}: The full page text is classified by a transformer-based text classifier to assign a document category (e.g., financial, legal, technical, scientific). This metadata enriches the JSON output and is included in all export formats.
\end{enumerate}

The parallel export makes sure that all formats are made with delay based on the slowest conversion, not the total. A single extraction run gives the user structured JSON, Markdown that people can read, document chunks that are ready for RAG, and a knowledge graph.

\subsubsection{Text binding.}

For each element box, text is obtained as follows:
\begin{itemize}
  \item For text-like elements (title, section, text, list\_item, headers, footers, etc.): text is extracted from the PDF \texttt{textpage} using the box in PDF coordinates (with y-axis inversion as needed).
  \item For \texttt{table} and \texttt{image}: the box is cropped from the page image, converted to base64, and passed to \textbf{EasyOCR} (English); the OCR result is stored as the element’s text. This ensures tables and figures have a text representation even when the PDF has no text layer.
\end{itemize}

A title-specific normalizer cleans up the title and section text (for example, it gets rid of non-UTF-8 characters and too much punctuation). The body text is normalized (for example, NFKC, bullet, and whitespace normalization) to make it more consistent and easier to match. Each detection is turned into a \textbf{Entity}: id (a unique UUID), type (a label), confidence, value (like \texttt{\{text: ...\}}), pixel coordinates (left, top, right, and bottom), mid-point, x\_center, y\_center, weight (from a fixed schema used for ordering and graph edges), and for table/image a \texttt{base64\_image} for optional LLM processing. Entities with no text or extremely little text (such $<$ 3 characters) are only kept for table and image types. This way, the layout structure stays the same even when PDF text is missing.

\subsubsection{Image classifier gate and Vision LLM insights.}

This is a crucial aspect of the pipeline: after pieces are found, each image or figure is first put through an image classifier (ViT) that tells it whether it is valuable or not. Only images that are deemed relevant are taken into account for downstream enrichment. Useless photos are stored in \texttt{skipped\_images} and are not included in either the Vision LLM or content exports (like Markdown and RAG papers). The ViT does not sort tables, and they are always kept. When \texttt{skip\_images} is enabled, the classifier runs on all image elements in parallel; when \texttt{skip\_insights} is false, every \emph{non-skipped} image and every table is sent to a \textbf{Vision LLM} (e.g., GPT-4o-mini) with a fixed prompt (from \texttt{insights/image.yaml}) that asks for \texttt{title}, \texttt{summary}, and \texttt{text} (string or list-of-dict for tables). So, the Vision LLM is only called for images that pass the usefulness gate (and for all tables). This cuts down on API costs and stops low-value captions from being used for ornamental or unimportant figures. LLM runs are done on each page at the same time, and the resultant JSON is processed and added to the entity \texttt{value}.

\subsection{Layout Grouping and Reading Order}
\label{sec:grouping}

\subsubsection{Group assignment.}

For each layout bounding box (multi-column, row, or group), the pipeline finds all element boxes whose \emph{midpoint} lies inside the layout box (excluding \texttt{page\_header}, \texttt{page\_footer}, \texttt{table\_of\_content}). Those elements form a candidate set. Then:
\begin{itemize}
  \item \textbf{Multi-column} (\texttt{multi\_column\_text}): the candidates’ x\_centers are normalized (MinMaxScaler) and clustered with \textbf{DBSCAN} (eps=0.3, min\_samples=2). Each cluster is sorted by y\_center (descending) and turned into a \textbf{Group} of type \texttt{multi-col} with a combined bounding box.
  \item \textbf{Row group} (\texttt{row\_group}): candidates are sorted by (x\_center, -y\_center); adjacent pairs are checked with \texttt{line\_angle}. If the angle is $\ge 50^\circ$, the pair is swapped to respect visual row order. The ordered list becomes a \texttt{row} Group.
  \item \textbf{Generic group}: candidates are sorted by \texttt{y\_center} (reading order) and form a single \texttt{group} Group.
\end{itemize}

Each group keeps track of its kind, ids (ordered element IDs), pixel coordinates (the group's bounding box), mid point, x center, and y center. This representation can be serialized in reading order and changed to DP-Bench or other formats that demand items to be grouped or sorted.

\subsubsection{Page-level ordering.}

First, elements are sorted by $(-\texttt{y\_center}, \texttt{x\_center})$, from top to bottom and then from left to right. The function \texttt{get\_unique\_pages} removes duplicate detections (maintaining higher-confidence when content matches). Then, by vertical position, \texttt{order\_page\_elements} combines \texttt{non\_groups} (elements that aren't in any group, sorted by \texttt{top}) with \texttt{groups}. The \texttt{elements} dictionary is rebuilt in this order (Python 3.7+ insertion order). This gives each page a single, stable reading order.

\subsubsection{Header and footer correction.}

Fuzzy string matching (e.g., thefuzz ratio $(> 95$) is used to match up headers and footers across pages. First, candidate headers and footers are gathered from elements that have previously been marked as such. Then, the text of every other element is compared to these candidates. The type of a text block is set to page\_header or page\_footer if it matches a known header or footer string on another page. Position heuristics (such top > 100 vs. bottom in pixel space) can switch the wrong header/footer when the detector becomes confused about the top and bottom areas. This step makes NID and RAG better by making sure that repetitive boilerplate is always labeled and can be removed during conversion if needed.

\subsection{Output Formats}

The extraction result is a JSON structure: \texttt{filename}, \texttt{total\_pages}, \texttt{total\_processed\_pages}, \texttt{total\_failed\_pages}, \texttt{total\_llm\_calls}, \texttt{metadata}, \texttt{document\_category}, and \texttt{pages}. Each page has \texttt{page\_number}, \texttt{groups}, \texttt{non\_groups}, \texttt{skipped\_images}, and \texttt{elements} (dict of id $\to$ entity, in reading order).

From this JSON, three conversions run \textbf{in parallel} (again via \texttt{ThreadPoolExecutor}); the data flow is shown in Listing~

\begin{enumerate}
  \item \textbf{Markdown}: For each page, elements are iterated in order. Titles become \texttt{\#\#}, sections \texttt{\#\#\#}, text as paragraphs, list items as bullets (with \texttt{extract\_list\_items}), tables/images with optional caption/summary and table data as Markdown tables (pandas \texttt{to\_markdown}). Headers/footers can be skipped or emitted as annotated blocks.
  \item \textbf{Documents (RAG)}: Three kinds of chunks are produced: (1) one page-level document (all element texts concatenated), (2) header-based blocks (section/title plus following list\_item/text), and (3) one document per element. Each has \texttt{page\_content} and metadata (page number, type, token count, filename, document\_category, etc.). Duplicates are removed by content hash.
  \item \textbf{Knowledge graph}: A graph is built with a root node (filename), page nodes, and one node per element (in reading order). Edges: root $\to$ page (\texttt{contains}), page $\to$ first element (\texttt{contains}), and between consecutive elements (\texttt{sibling} if same weight, \texttt{parent-child} otherwise). Weights come from the schema (e.g., title=1, section=2, table=3, text=6, footer=7).
\end{enumerate}

\subsection{Data and Schema}

The schema defines the types and weights of elements. For example, the schema defines the types and weights of entities as follows: \texttt(Entity (id, type, confidence, value, pixel\_coordinates, mid\_point, x\_center, y\_center, weight, base64\_image) and Group (type, ids, pixel\_coordinates, mid\_point, x\_center, y\_center)). The "value" field contains "text" and, for tables and images, optional "title," "summary," and structured "text" from the LLM (for example, a list of dictionaries for tabular data). Weights (such title=1, section=2, table=3, text=6, and footer=7) control both the meaning of the reading order and the labels for knowledge graph edges (sibling vs. parent-child). This schema is the only place to find out how to order things and how to change them to DP-Bench format (category mapping and changing coordinates from left/top/right/bottom to four-point polygons).

\section{Evaluation on DP-Bench}
\label{sec:eval}

\subsection{Benchmark: upstage/dp-bench}

We use the DP-Bench dataset, which you can get on Hugging Face at upstage/dp-bench. It gives you reference annotations for processing JSON documents. Each sample has elements with coordinates, category, id, page, and content (text, html, or markdown). There are different types of categories, such as Table, Paragraph, Figure, Chart, Header, Footer, Caption, Equation, Heading1, List, Index, and Footnote. There are 90 samples from the Library of Congress, 90 from Open Educational Resources, and 20 from Upstage in the collection. These samples are from a variety of fields, including social sciences, natural sciences, technology, and mathematics. The benchmark is a good example of how real-world documents mix because the layout elements are not balanced (for example, there are a lot of paragraphs and headings but not many tables and equations). DP-Bench's evaluation scripts use reference and prediction JSON files to figure out NID (layout mode) and TEDS/TEDS-S (table mode).

\subsection{Metrics}

\begin{itemize}
  \item \textbf{NID (Normalized Indel Distance)}: Measures how well the parser detects and serializes document elements in human reading order. Defined as
  \[
  \text{NID} = 1 - \frac{\text{distance}}{\text{len(reference)} + \text{len(prediction)}},
  \]
  where the distance is character-level insertions/deletions (no substitutions). Higher NID means better layout and ordering. Tables, figures, and charts are excluded from NID.
  \item \textbf{TEDS (Tree Edit Distance-based Similarity)}: For tables only. Compares predicted and reference tables as trees (structure + cell content):
  \[
  \text{TEDS}(T_a, T_b) = 1 - \frac{\text{EditDist}(T_a, T_b)}{\max(|T_a|, |T_b|)}.
  \]
  \item \textbf{TEDS-S}: Same formula but with structure-only trees (cell content omitted), measuring structural table recognition. DP-Bench reports both TEDS and TEDS-S for table-only evaluation.
\end{itemize}

\subsection{Mapping NovaLAD to DP-Bench}

The following are the NovaLAD element types and their corresponding DP-Bench categories: \texttt{page\_header} $\to$ Header, \texttt{text} $\to$ Paragraph, \texttt{section} / \texttt{title} $\to$ Heading1, \texttt{list\_item} $\to$ List, \texttt{page\_footer} $\to$ Footer, \texttt{table\_caption} / \texttt{image\_caption} $\to$ Caption, \texttt{table\_of\_content} $\to$ Paragraph, \texttt{image} $\to$ Figure, \texttt{table} $\to$ Table. The coordinates are changed from NovaLAD's "left, top, right, bottom" to DP-Bench's list of four points: "left, top," "right, top," "right, bottom," and "left, bottom." The ordered \texttt{elements} dict is serialized in the order that NID expects. You can transfer Equation, Chart, Index, and Footnote from enhanced NovaLAD types or leave them for future schema additions. The existing mapping covers most of the DP-Bench categories.

\subsection{Model Training and Evaluation}

The NovaLAD pipeline uses two YOLO object detection models that work at the same time: Layout Detection and Element Detection. Both models were trained from the ground up utilizing bespoke datasets. This part shows the full training results, loss curves, and performance metrics for each model. Some important metrics are accuracy, recall, mAP50 (mean average precision at IoU 0.5), and mAP50-95 (mean average precision at IoU 0.5--0.95).

For detection, let TP, FP, and FN denote the number of true positives, false positives, and false negatives at an IoU threshold $\ge$. Precision and recall are defined as
\[
\text{Precision} = \frac{\text{TP}}{\text{TP} + \text{FP}}, \qquad
\text{Recall} = \frac{\text{TP}}{\text{TP} + \text{FN}},
\]
where a detection is a true positive if its predicted bounding box overlaps a ground-truth box with an intersection-over-union (IoU) value of at least $\tau$. The average precision (AP) is the area under the precision-recall curve for each class. To get the curve, you have to change the confidence level. mAP50 is the average AP across classes at $\ge=0.5$, while mAP50--95 is the average AP over $\ge \in \{0.5,0.55,\ldots,0.95\}$. This gives a more rigorous measure of localization quality.

YOLO optimizes three loss terms during training: box regression loss (how close the predicted box is to the ground truth), classification loss (the chance that the right class is assigned), and distribution focal loss (DFL, which improves the discrete bounding-box distribution). Objectness losses show how well the model can tell if a place has any objects at all. We report accuracy for the image classifier.
\[
\text{Accuracy} = \frac{\text{number of correct predictions}}{\text{total number of predictions}},
\]
and F1-score,
\[
\text{F1} = 2 \cdot \frac{\text{Precision} \cdot \text{Recall}}{\text{Precision} + \text{Recall}},
\]
which balances precision and recall on the \emph{Useful} vs \emph{Useless} classes.

\subsubsection{Layout Detection Model}
\label{sec:eval-layout}

The Layout Detection model was trained for 30 epochs to find structural areas like layout\_box, column\_group, column\_text, group, multi\_column, and row\_group. Table~\ref{tab:layout-training} shows the final training metrics in a summary.

\begin{table}[t]
\centering
\caption{Layout detection YOLO: final training metrics (epoch 30). Detects layout\_box, column\_group, multi\_column, row\_group, etc.}
\label{tab:layout-training}
\small
\begin{tabular}{lc}
\toprule
Metric & Value \\
\midrule
Precision & 0.619 \\
Recall & 0.530 \\
mAP50 & 0.567 \\
mAP50-95 & 0.512 \\
\midrule
Epochs & 30 \\
\bottomrule
\end{tabular}
\end{table}

\begin{figure*}[t]
\centering
\begin{subfigure}[b]{0.32\textwidth}
  \centering
  \includegraphics[width=\linewidth]{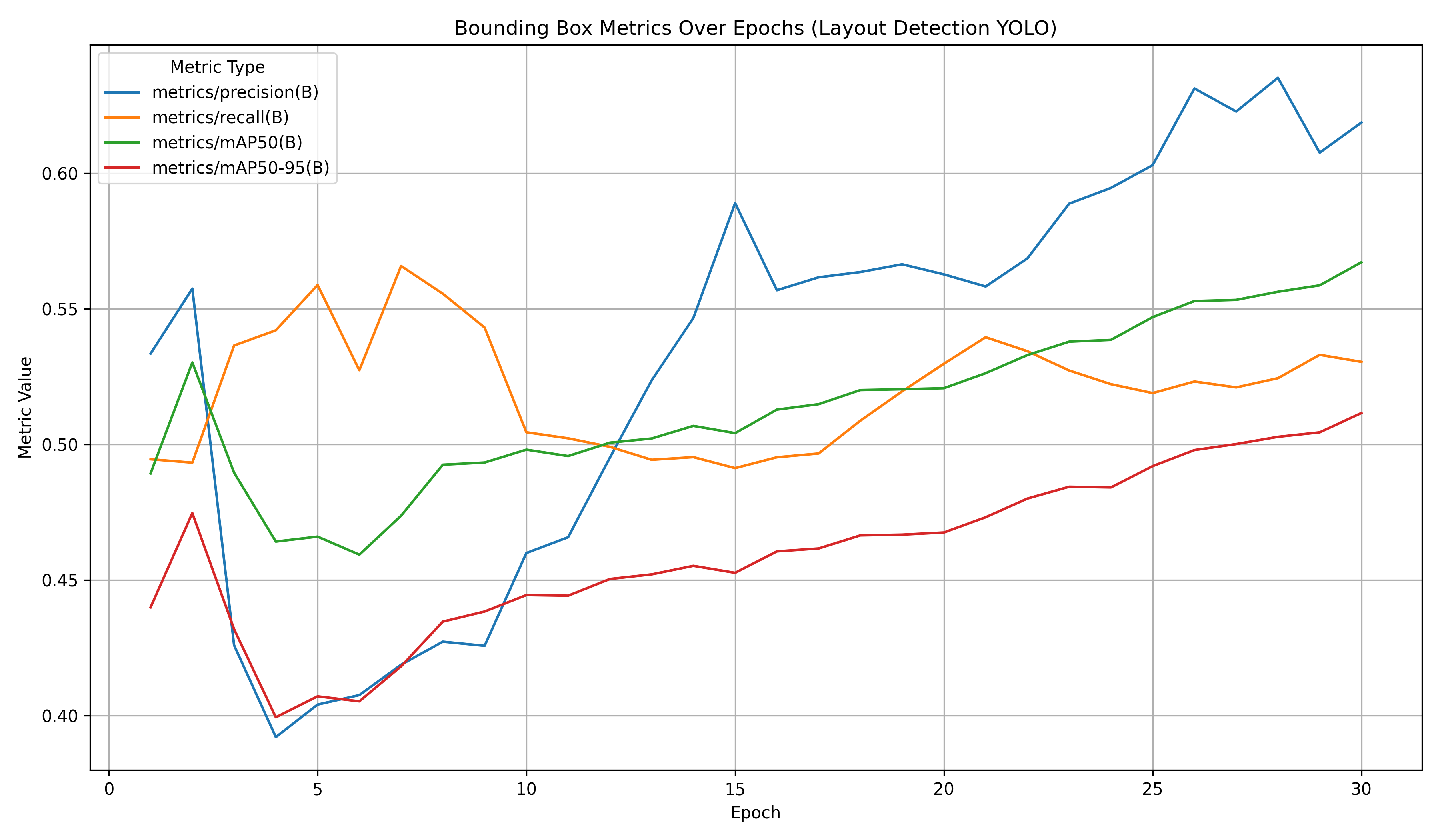}
  \caption{Precision/recall and mAP.}
\end{subfigure}
\begin{subfigure}[b]{0.32\textwidth}
  \centering
  \includegraphics[width=\linewidth]{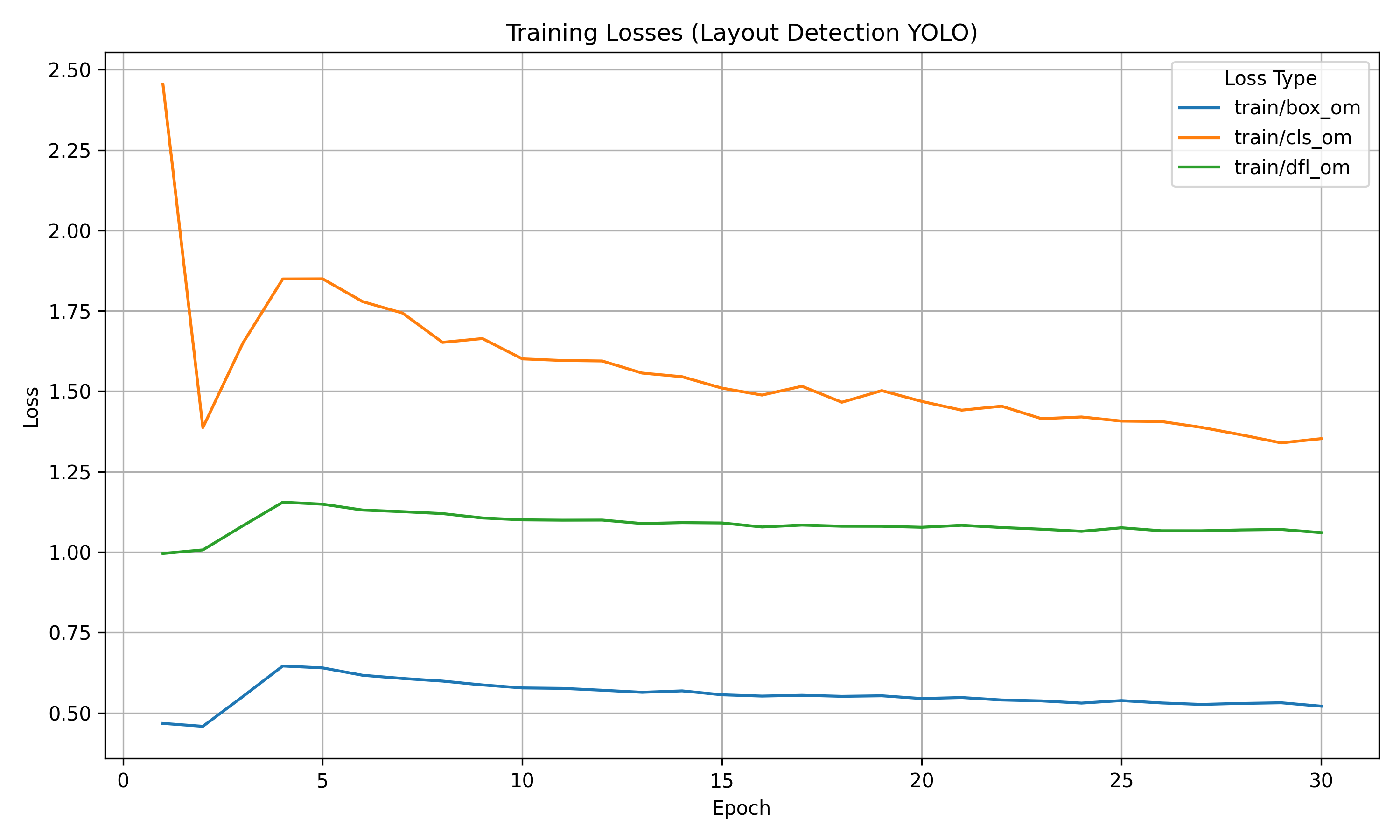}
  \caption{Training loss components.}
\end{subfigure}
\begin{subfigure}[b]{0.32\textwidth}
  \centering
  \includegraphics[width=\linewidth]{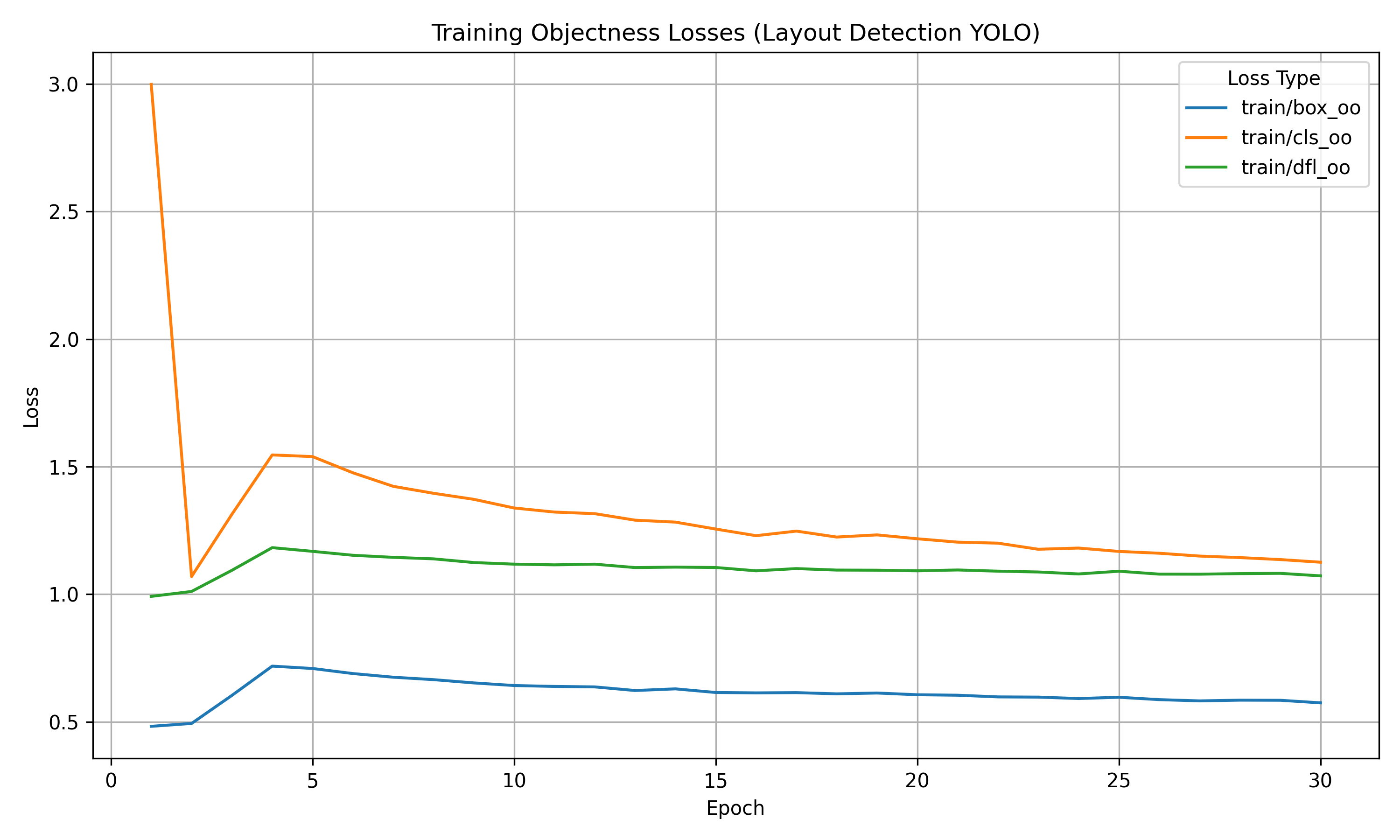}
  \caption{Training objectness losses.}
\end{subfigure}

\vspace{0.5em}
\begin{subfigure}[b]{0.32\textwidth}
  \centering
  \includegraphics[width=\linewidth]{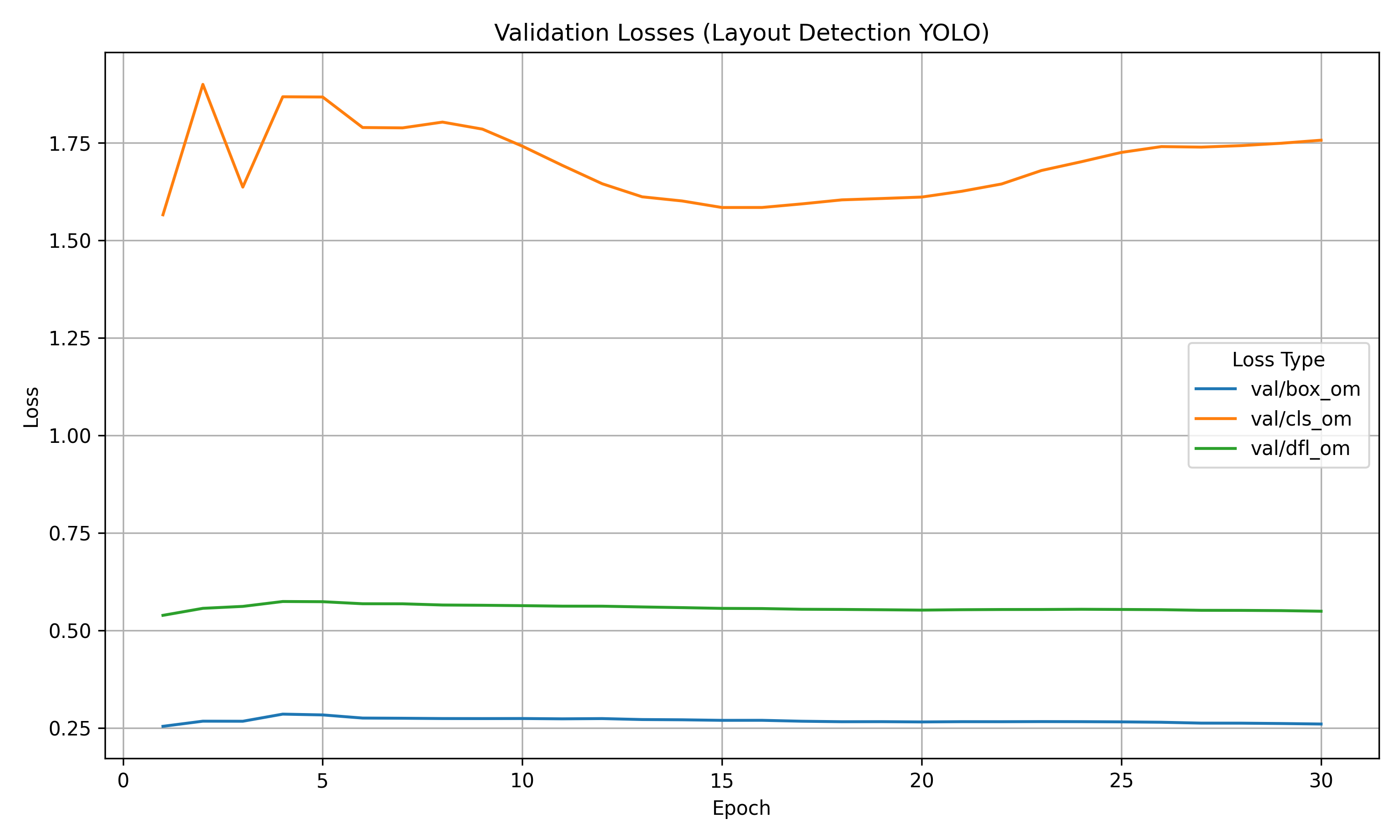}
  \caption{Validation loss components.}
\end{subfigure}
\begin{subfigure}[b]{0.32\textwidth}
  \centering
  \includegraphics[width=\linewidth]{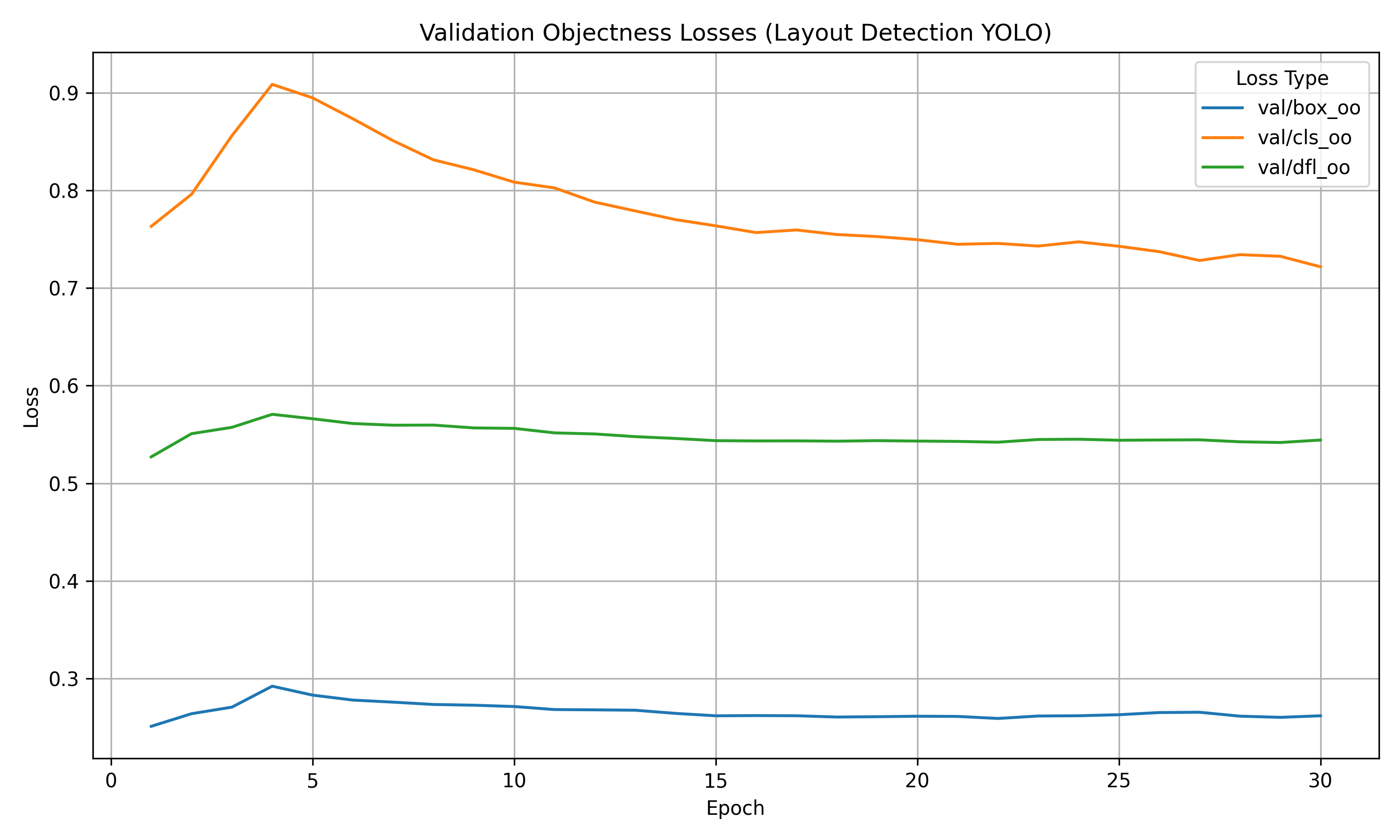}
  \caption{Validation objectness losses.}
\end{subfigure}
\begin{subfigure}[b]{0.32\textwidth}
  \centering
  \includegraphics[width=\linewidth]{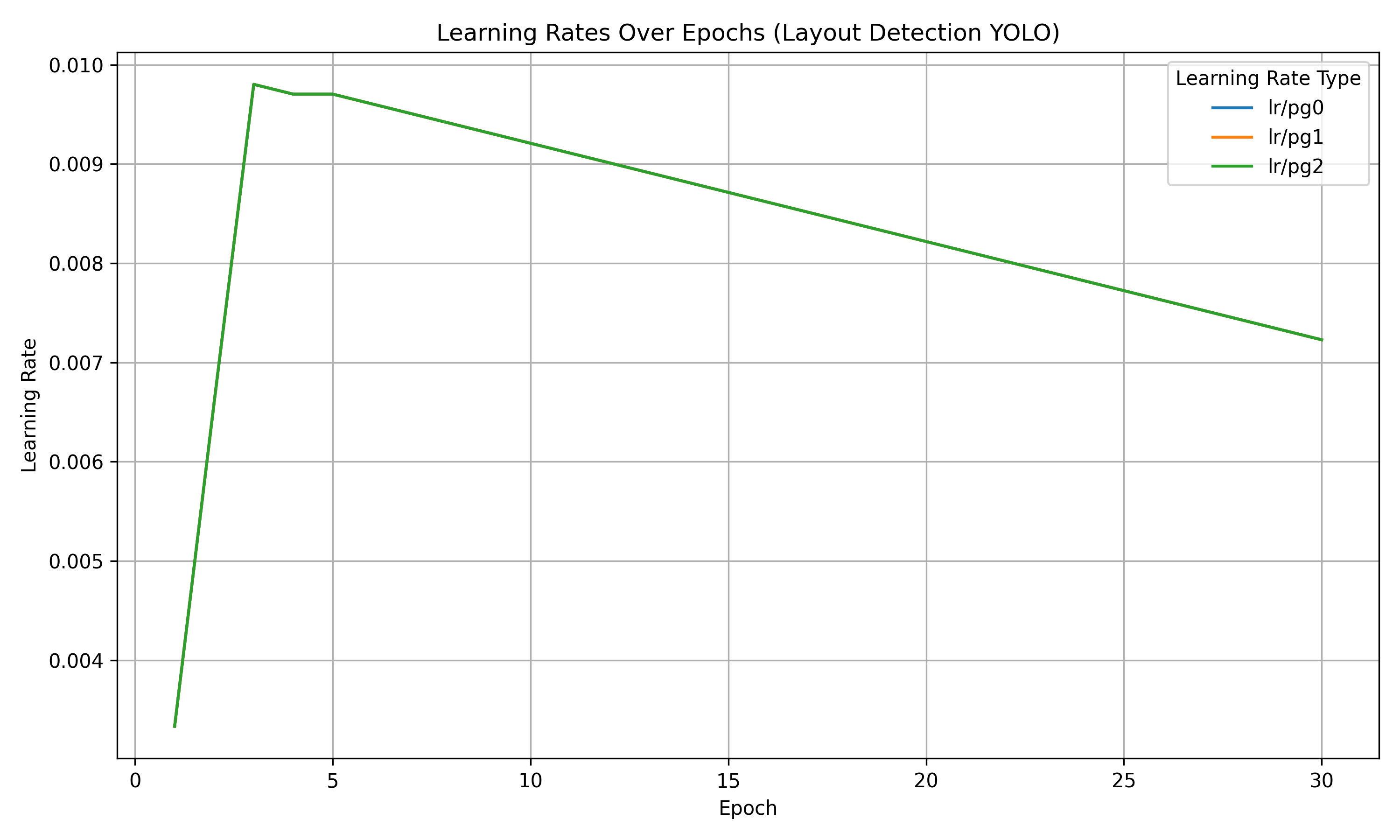}
  \caption{Learning-rate schedule.}
\end{subfigure}
\caption{Training and validation dynamics for the layout detection YOLO model. (a) shows how precision, recall, mAP50, and mAP50--95 improve over 30 epochs; (b)--(e) decompose the box, classification, DFL, and objectness losses on train and validation splits; (f) visualizes the learning-rate schedule used during training.}
\label{fig:layout-training-overview}
\end{figure*}

Figure~\ref{fig:layout-training-overview} illustrates that layout detection is the harder of the two YOLO tasks: precision and recall increase steadily but saturate below the element detector due to visually subtle boundaries between regions such as column\_group and row\_group. The box loss stabilizes early, indicating that the model localizes regions reliably, while the larger classification loss reflects the difficulty of assigning the correct layout type. The close match between training and validation loss curves suggests limited overfitting and good generalization to unseen page layouts.

\subsubsection{Element Detection Model}
\label{sec:eval-elements}

The Element Detection model was trained for 50 epochs to detect semantic document elements: title, header, page\_header, text, table, image, list\_item, and captions. Table~\ref{tab:element-training} summarizes the final training metrics.

\begin{table}[H]
\centering
\caption{Element detection YOLO: final training metrics (epoch 50). Detects title, header, page\_header, text, table, image, etc.}
\label{tab:element-training}
\small
\begin{tabular}{lc}
\toprule
Metric & Value \\
\midrule
Precision & 0.806 \\
Recall & 0.813 \\
mAP50 & 0.859 \\
mAP50-95 & 0.670 \\
\midrule
Epochs & 50 \\
\bottomrule
\end{tabular}
\end{table}

\begin{figure*}[t]
\centering
\begin{subfigure}[b]{0.32\textwidth}
  \centering
  \includegraphics[width=\linewidth]{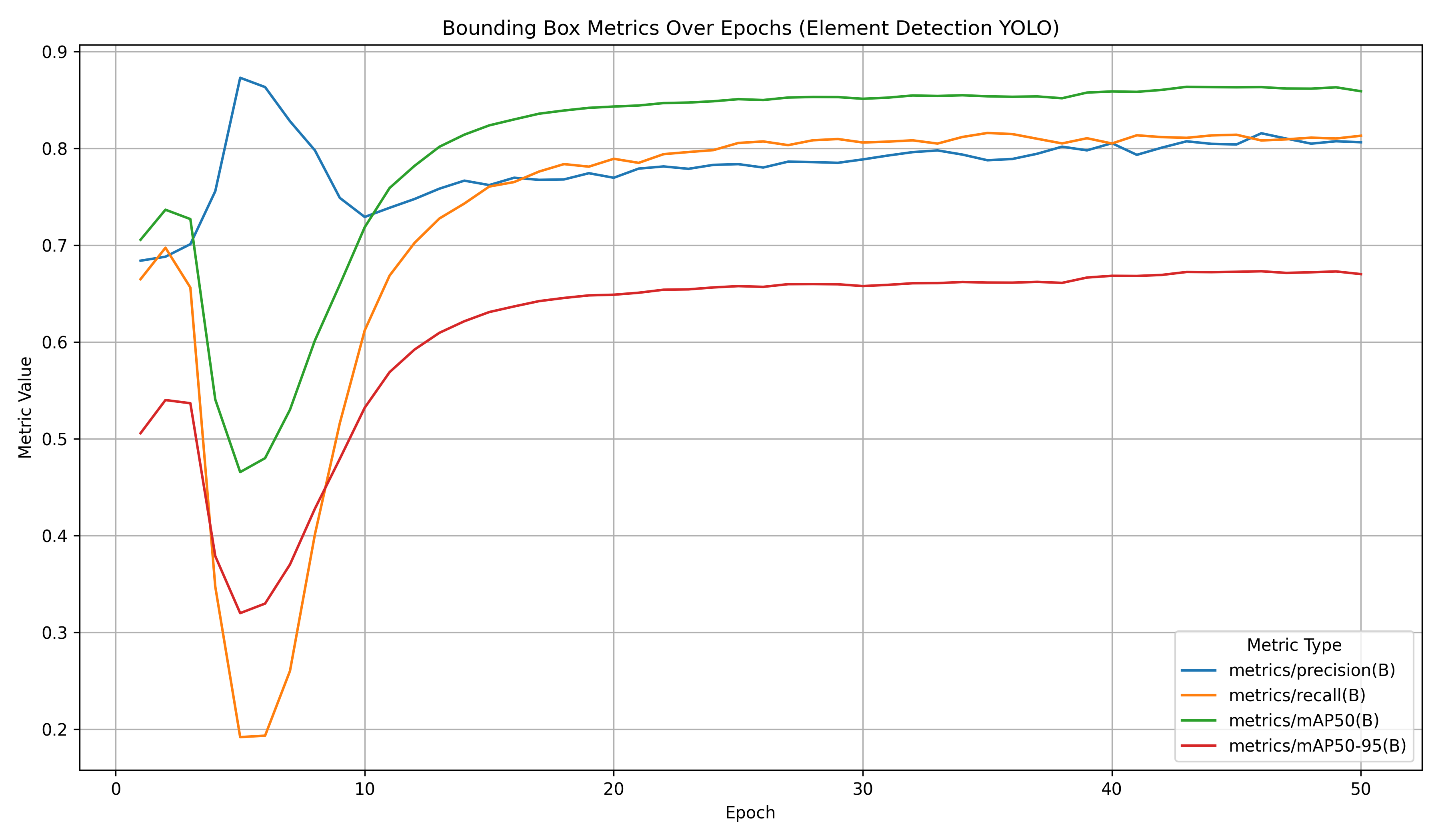}
  \caption{Precision/recall and mAP.}
\end{subfigure}
\begin{subfigure}[b]{0.32\textwidth}
  \centering
  \includegraphics[width=\linewidth]{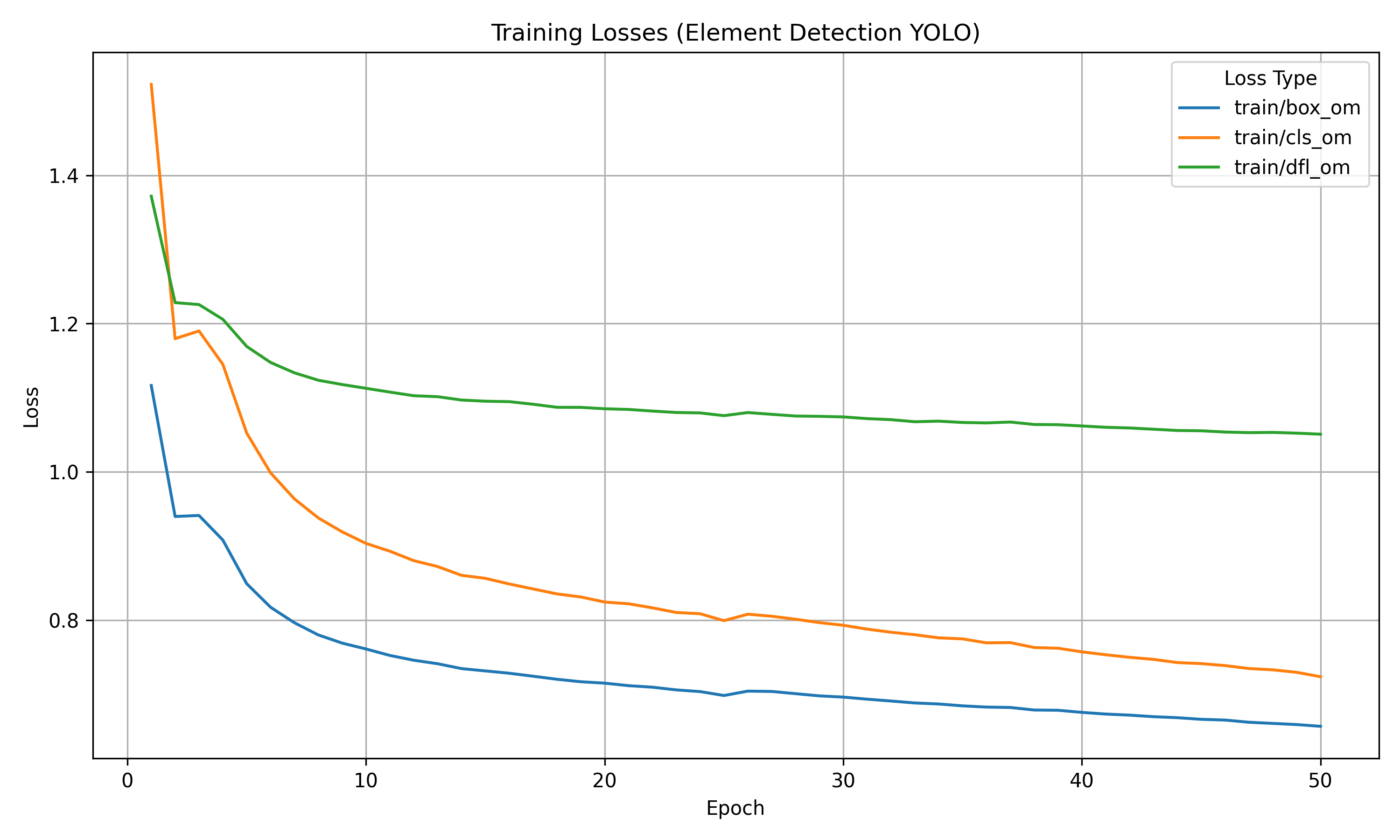}
  \caption{Training loss components.}
\end{subfigure}
\begin{subfigure}[b]{0.32\textwidth}
  \centering
  \includegraphics[width=\linewidth]{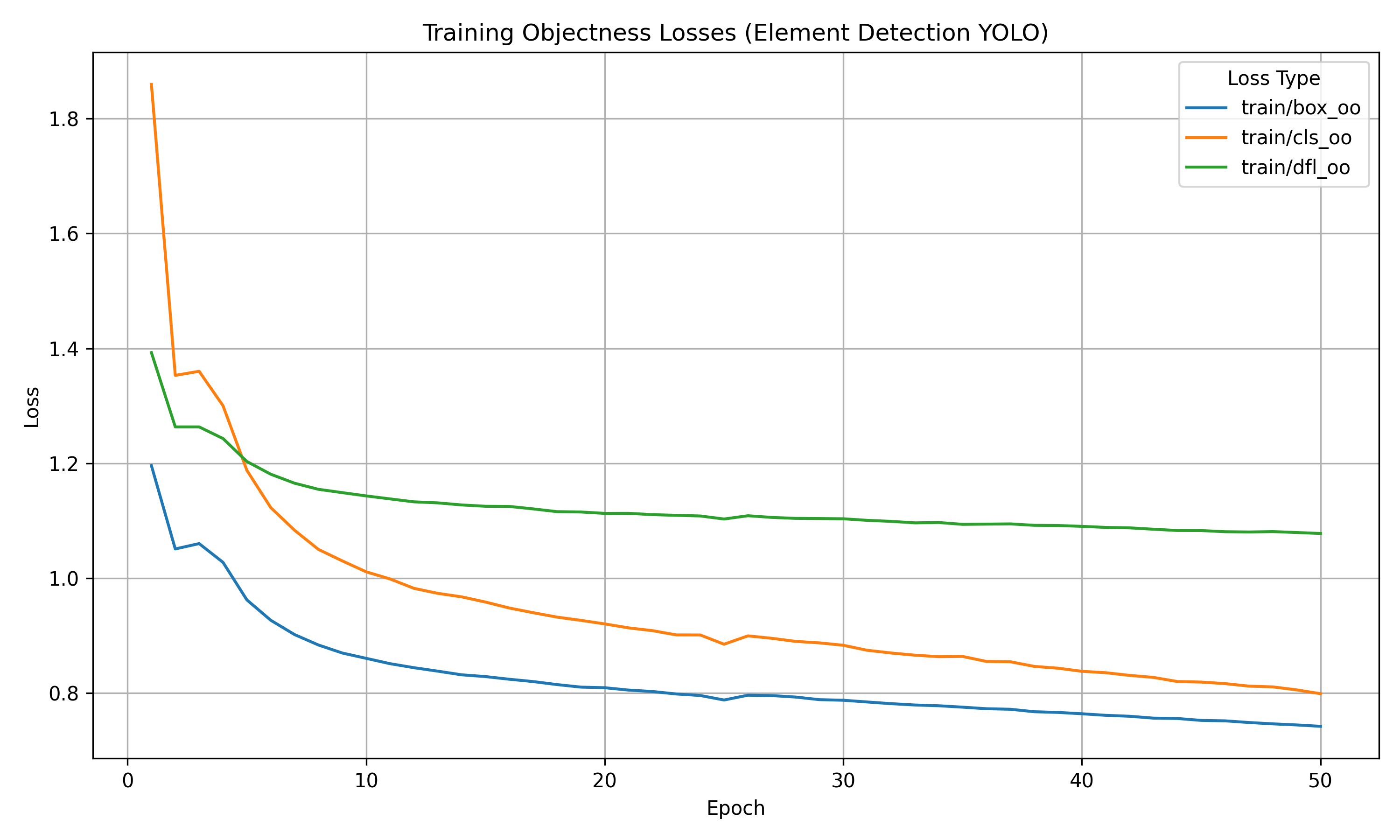}
  \caption{Training objectness losses.}
\end{subfigure}

\vspace{0.5em}
\begin{subfigure}[b]{0.32\textwidth}
  \centering
  \includegraphics[width=\linewidth]{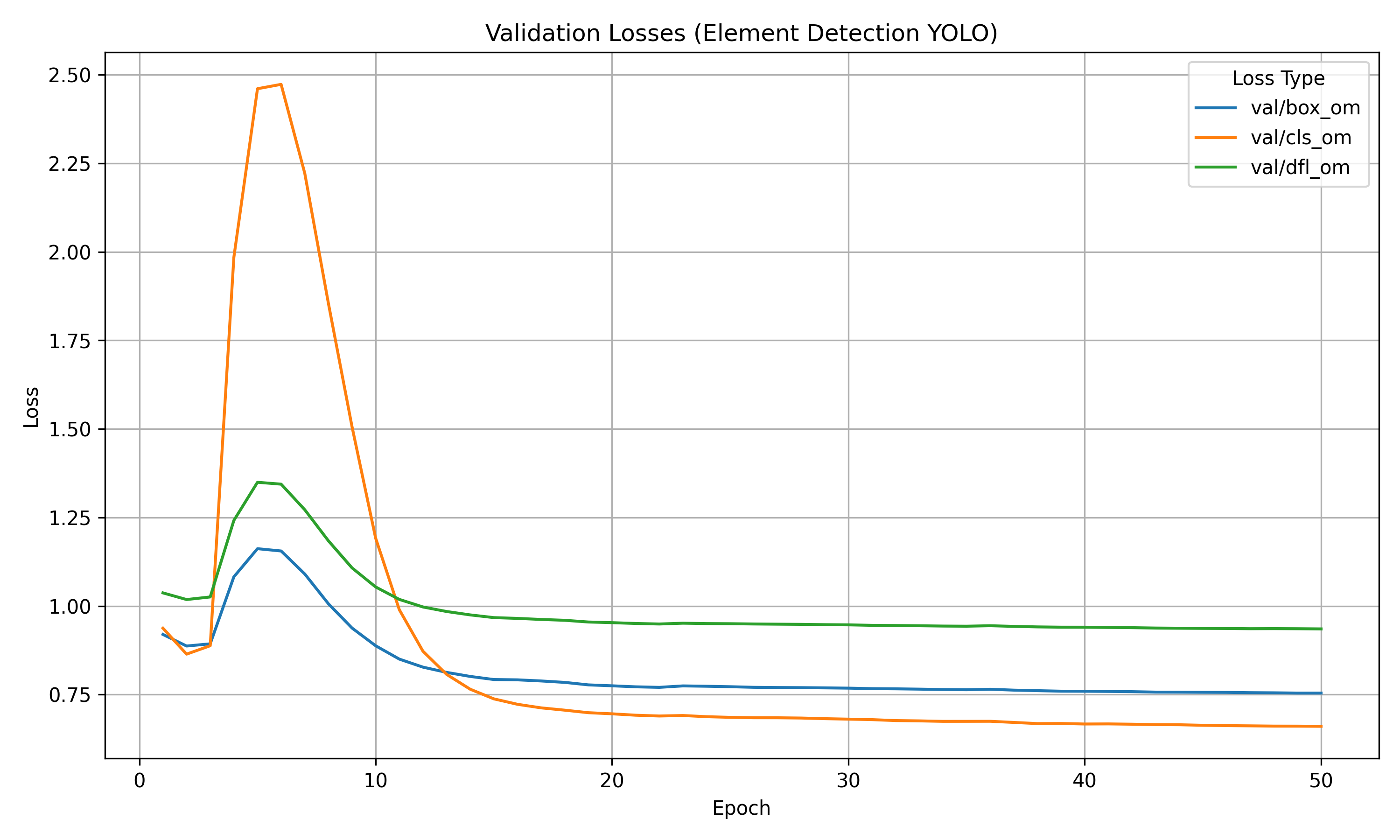}
  \caption{Validation loss components.}
\end{subfigure}
\begin{subfigure}[b]{0.32\textwidth}
  \centering
  \includegraphics[width=\linewidth]{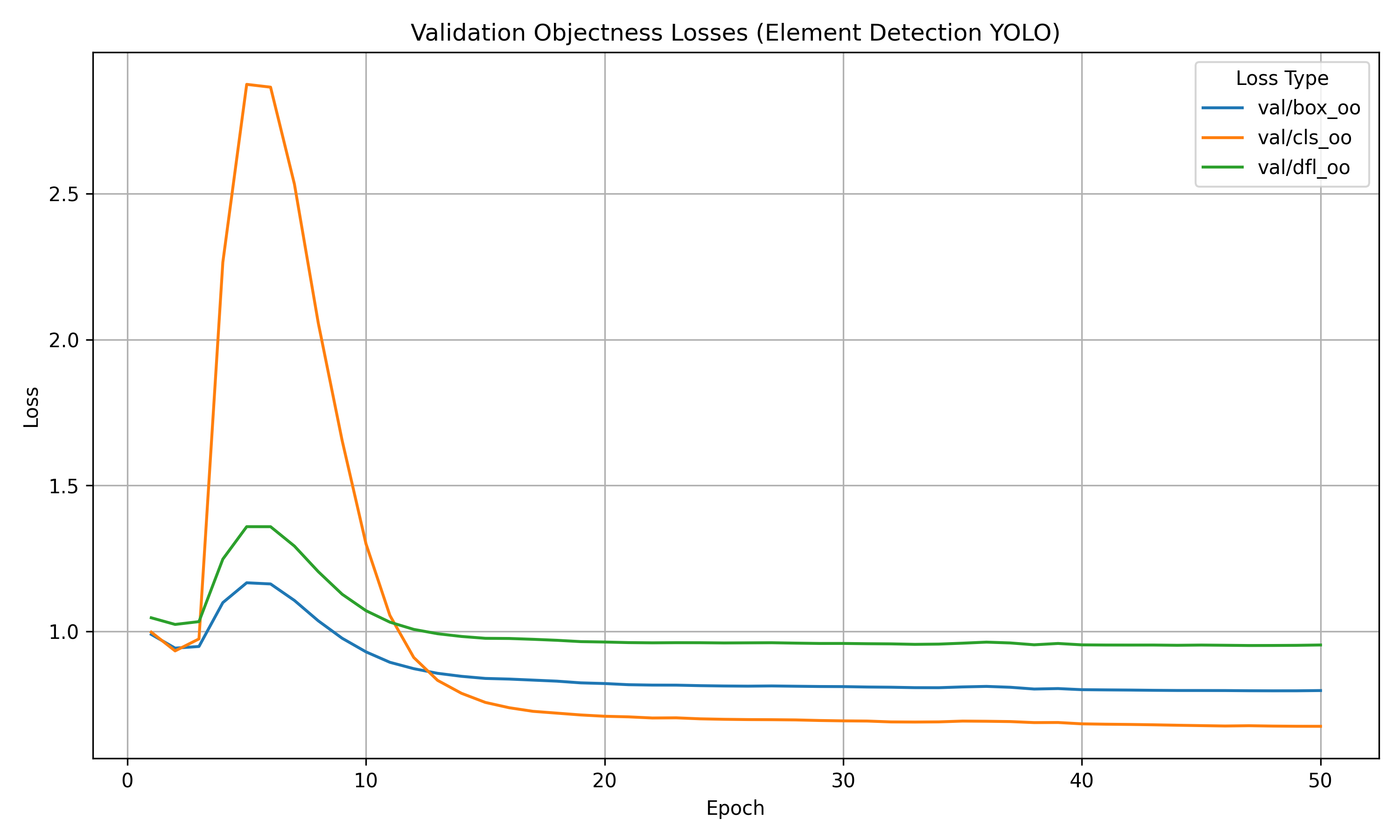}
  \caption{Validation objectness losses.}
\end{subfigure}
\begin{subfigure}[b]{0.32\textwidth}
  \centering
  \includegraphics[width=\linewidth]{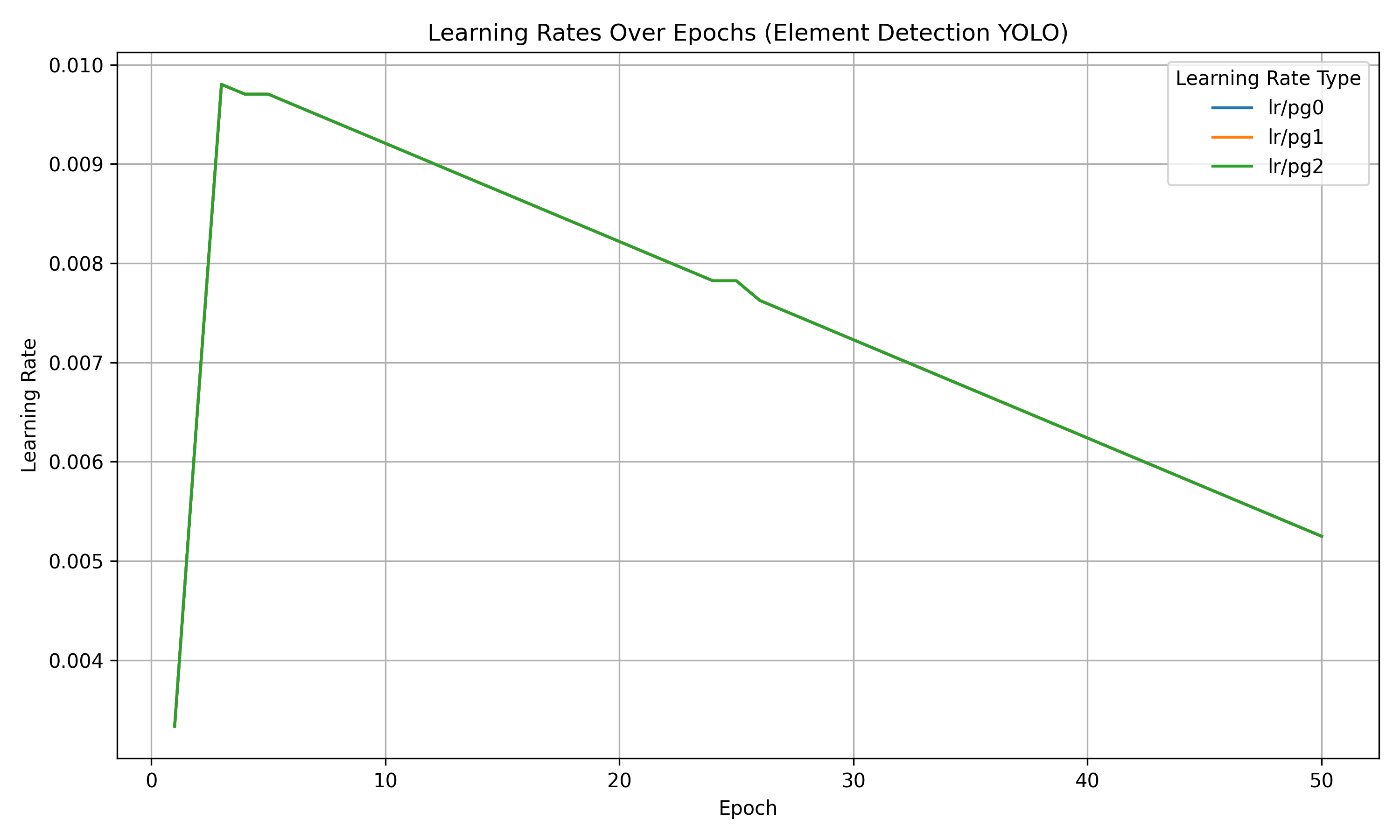}
  \caption{Learning-rate schedule.}
\end{subfigure}
\caption{Training and validation dynamics for the element detection YOLO model. (a) shows precision, recall, mAP50, and mAP50--95 over 50 epochs; (b)--(e) detail how box, classification, DFL, and objectness losses evolve on train and validation sets; (f) shows the corresponding learning-rate schedule.}
\label{fig:element-training-overview}
\end{figure*}

Figure~\ref{fig:element-training-overview} shows that the element detector converges to substantially higher mAP than the layout model, with mAP50 reaching $\sim$0.86 by epoch~50. All loss components decline smoothly, and the validation curves track the training curves closely, which indicates that the model is neither underfitting nor overfitting. The stronger performance reflects that semantic elements (titles, text blocks, tables, images) have more distinctive visual signatures than abstract layout regions, which NovaLAD exploits to drive high-quality downstream extraction.

Both models demonstrate consistent convergence and efficient training, validating the design decision of parallel dual detection in the NovaLAD pipeline. The element model gets better final metrics because semantic elements have more distinguishing visual properties than layout regions.

\subsubsection{Image Classifier (ViT) Model}
\label{sec:eval-image-classifier}

Before choosing the final model, we looked at a lot of different combinations of backbone architectures and classifiers. Table~\ref{tab:classifier-comparison} shows how EfficientNetB0, ResNet50, CLIP (ViT-B/32), and ViT-Base (Patch16-224) compare to several classifiers, such as XGBoost, LightGBM, ExtraTrees, CatBoost, Linear, MLP, and Sigmoid. We chose ViT-Base (google/vit-base-patch16-224-in21k) fine-tuned end-to-end since it works better (98.53\% accuracy), takes up less space, and infers faster than bulkier ResNet50-based setups and CLIP-based feature extractors with separate classifiers. The ViT model was fine-tuned for binary classification: \emph{Useful} (pictures that have information, charts, flowcharts, diagrams) vs \emph{Useless} (decorative images, logos, or placeholders to skip). Table~\ref{tab:vit-results} shows the final fine-tuned results in a short form.

\begin{table*}[htbp] 
\centering
\caption{Performance comparison of different backbones and classifiers for the image usefulness task. Class 0: Useful; Class 1: Useless.}
\label{tab:classifier-comparison}
\resizebox{\textwidth}{!}{
\begin{tabular}{lllccccc} 
\hline
\textbf{Backbone} & \textbf{Classifier} & \textbf{Class} & \textbf{Precision} & \textbf{Recall} & \textbf{F1-score} & \textbf{Accuracy (\%)} & \textbf{Time (s)} \\ \hline
\multirow{8}{*}{EfficientNetB0} 
& XGBoost & 0 & 0.9702 & 0.9743 & 0.9722 & \multirow{2}{*}{97.22} & \multirow{2}{*}{17.31} \\
&         & 1 & 0.9742 & 0.9700 & 0.9721 &                        &                        \\ \cline{2-8}
& LightGBM & 0 & 0.9677 & 0.9815 & 0.9745 & \multirow{2}{*}{97.43} & \multirow{2}{*}{22.54} \\
&          & 1 & 0.9812 & 0.9672 & 0.9741 &                        &                        \\ \cline{2-8}
& ExtraTrees & 0 & 0.9611 & 0.9857 & 0.9732 & \multirow{2}{*}{97.29} & \multirow{2}{*}{28.13} \\
&            & 1 & 0.9854 & 0.9601 & 0.9725 &                        &                        \\ \cline{2-8}
& CatBoost & 0 & 0.9701 & 0.9729 & 0.9715 & \multirow{2}{*}{97.15} & \multirow{2}{*}{10.43} \\
&          & 1 & 0.9728 & 0.9700 & 0.9714 &                        &                        \\ \hline
\multirow{8}{*}{ResNet50} 
& XGBoost & 0 & 0.9691 & 0.9843 & 0.9766 & \multirow{2}{*}{97.14} & \multirow{2}{*}{108.63} \\
&         & 1 & 0.9840 & 0.9686 & 0.9762 &                        &                         \\ \cline{2-8}
& LightGBM & 0 & 0.9705 & 0.9857 & 0.9780 & \multirow{2}{*}{97.57} & \multirow{2}{*}{3.95}   \\
&          & 1 & 0.9855 & 0.9700 & 0.9777 &                        &                         \\ \cline{2-8}
& ExtraTrees & 0 & 0.9504 & 0.9857 & 0.9677 & \multirow{2}{*}{97.29} & \multirow{2}{*}{6.33}   \\
&            & 1 & 0.9851 & 0.9486 & 0.9665 &                        &                         \\ \cline{2-8}
& CatBoost & 0 & 0.9691 & 0.9843 & 0.9766 & \multirow{2}{*}{97.64} & \multirow{2}{*}{26.65}  \\
&          & 1 & 0.9840 & 0.9686 & 0.9762 &                        &                         \\ \hline
\multirow{10}{*}{CLIP (ViT-B/32)} 
& Linear & 0 & 0.9912 & 0.9883 & 0.9897 & \multirow{2}{*}{99.00} & \multirow{2}{*}{2.14} \\
&        & 1 & 0.9888 & 0.9916 & 0.9902 &                        &                       \\ \cline{2-8}
& LightGBM & 0 & 0.9719 & 0.9900 & 0.9809 & \multirow{2}{*}{97.57} & \multirow{2}{*}{5.04} \\
&          & 1 & 0.9898 & 0.9714 & 0.9805 &                        &                       \\ \cline{2-8}
& MLP & 0 & 0.9942 & 0.9914 & 0.9928 & \multirow{2}{*}{99.20} & \multirow{2}{*}{16.84}    \\
&     & 1 & 0.9914 & 0.9943 & 0.9928 &                        &                          \\ \cline{2-8}
& CatBoost & 0 & 0.9747 & 0.9928 & 0.9837 & \multirow{2}{*}{97.64} & \multirow{2}{*}{18.98} \\
&          & 1 & 0.9927 & 0.9743 & 0.9834 &                        &                        \\ \cline{2-8}
& Sigmoid & 0 & 0.9957 & 0.9915 & 0.9936 & \multirow{2}{*}{99.35} & \multirow{2}{*}{5.33}  \\
&         & 1 & 0.9913 & 0.9956 & 0.9935 &                        &                        \\ \hline
\multirow{10}{*}{ViT-Base} 
& XGBoost & 0 & 0.9800 & 0.9800 & 0.9800 & \multirow{2}{*}{98.36} & \multirow{2}{*}{206.61} \\
&         & 1 & 0.9800 & 0.9800 & 0.9800 &                        &                         \\ \cline{2-8}
& CatBoost & 0 & 0.9800 & 0.9800 & 0.9800 & \multirow{2}{*}{98.29} & \multirow{2}{*}{80.28}  \\
&          & 1 & 0.9800 & 0.9800 & 0.9800 &                        &                         \\ \cline{2-8}
& LightGBM & 0 & 0.9815 & 0.9829 & 0.9822 & \multirow{2}{*}{98.21} & \multirow{2}{*}{79.81}  \\
&          & 1 & 0.9828 & 0.9814 & 0.9821 &                        &                         \\ \cline{2-8}
& MLP & 0 & 0.9829 & 0.9829 & 0.9829 & \multirow{2}{*}{98.29} & \multirow{2}{*}{8.37}     \\
&     & 1 & 0.9828 & 0.9828 & 0.9828 &                        &                           \\ \cline{2-8}
& Sigmoid & 0 & 0.9883 & 0.9869 & 0.9876 & \multirow{2}{*}{98.78} & \multirow{2}{*}{4.48}   \\
&         & 1 & 0.9873 & 0.9887 & 0.9880 &                        &                         \\ \hline
\end{tabular}
}
\end{table*}

\begin{table}[H]
\centering
\caption{Image classifier (ViT): final training and evaluation results. Binary classification: Useful vs Useless.}
\label{tab:vit-results}
\small
\begin{tabular}{lc}
\toprule
Metric & Value \\
\midrule
\multicolumn{2}{l}{\textit{Evaluation}} \\
Accuracy & 98.53\% \\
Loss & 0.0851 \\
\midrule
\multicolumn{2}{l}{\textit{Training}} \\
Train loss & 0.2049 \\
Epochs & 4 \\
\midrule
\multicolumn{2}{l}{\textit{Runtime}} \\
Train runtime & $\sim$7.9 h \\
Eval runtime & 45.2 s \\
Eval samples/s & 12.09 \\
\bottomrule
\end{tabular}
\end{table}

The model gets 98.53\% of the evaluations right, with an evaluation loss of 0.0851. After four epochs, the training loss reached 0.2049. The evaluation throughput is 12.09 samples per second, which makes it possible to quickly classify image elements in parallel within the pipeline. The high accuracy makes sure that important images (such charts, flowcharts, and diagrams) are kept for LLM enrichment and that unnecessary ornamental images are quickly filtered out, which cuts down on noise and costs later on.

\begin{figure*}[t]
\centering
\begin{subfigure}[b]{0.32\textwidth}
  \centering
  \includegraphics[width=\linewidth]{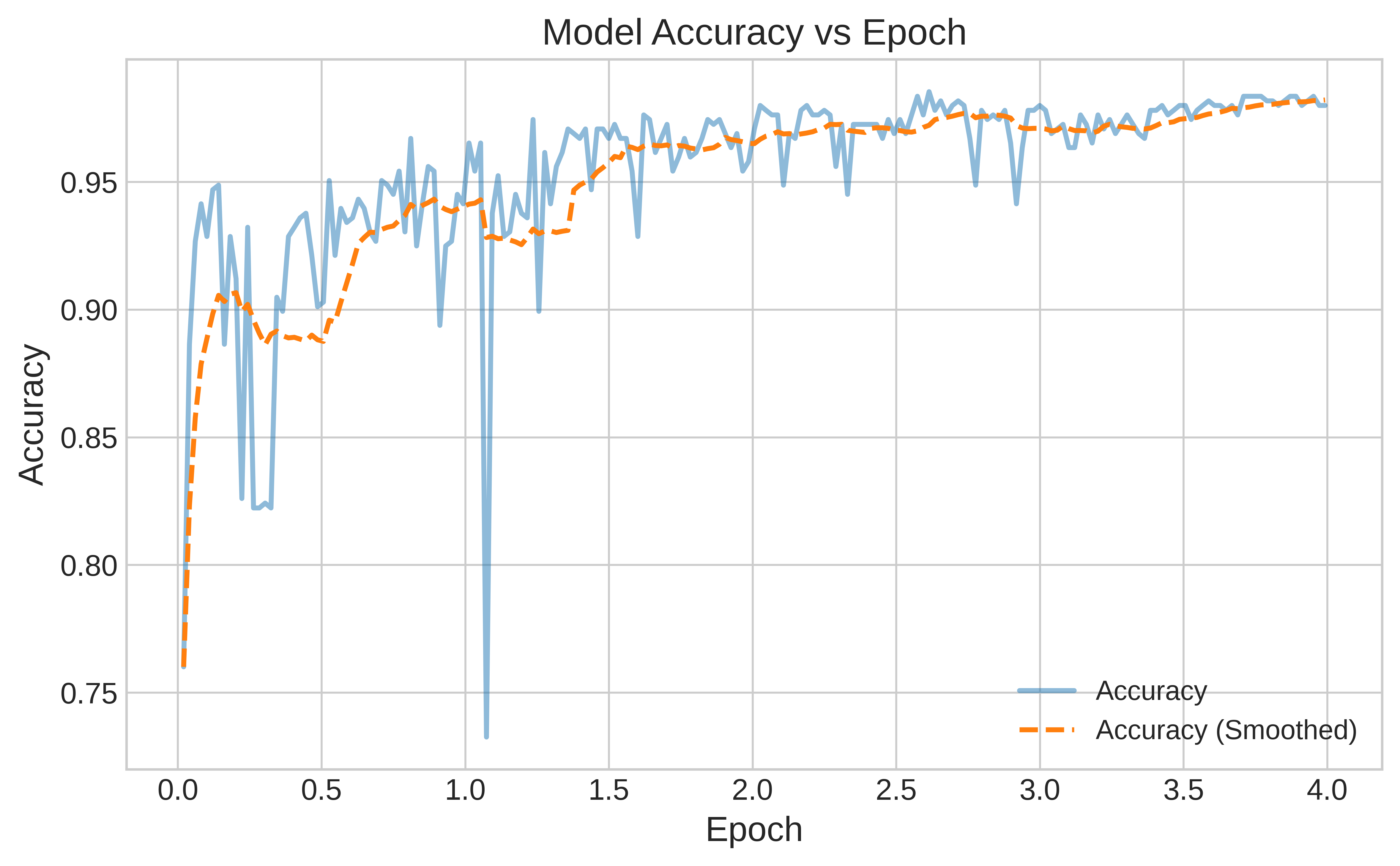}
  \caption{Evaluation accuracy vs.\ epoch.}
\end{subfigure}
\begin{subfigure}[b]{0.32\textwidth}
  \centering
  \includegraphics[width=\linewidth]{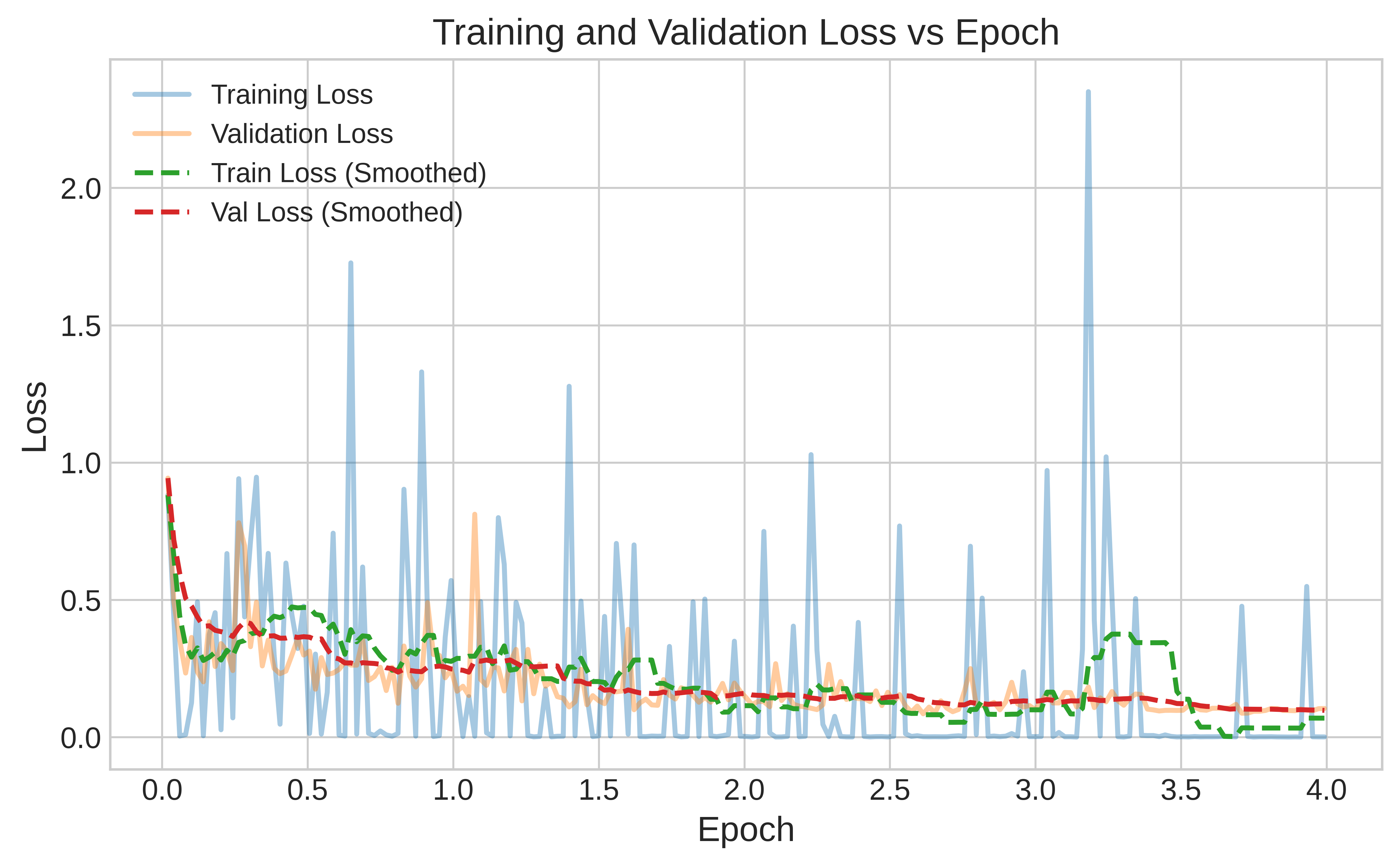}
  \caption{Training loss vs.\ epoch.}
\end{subfigure}
\begin{subfigure}[b]{0.32\textwidth}
  \centering
  \includegraphics[width=\linewidth]{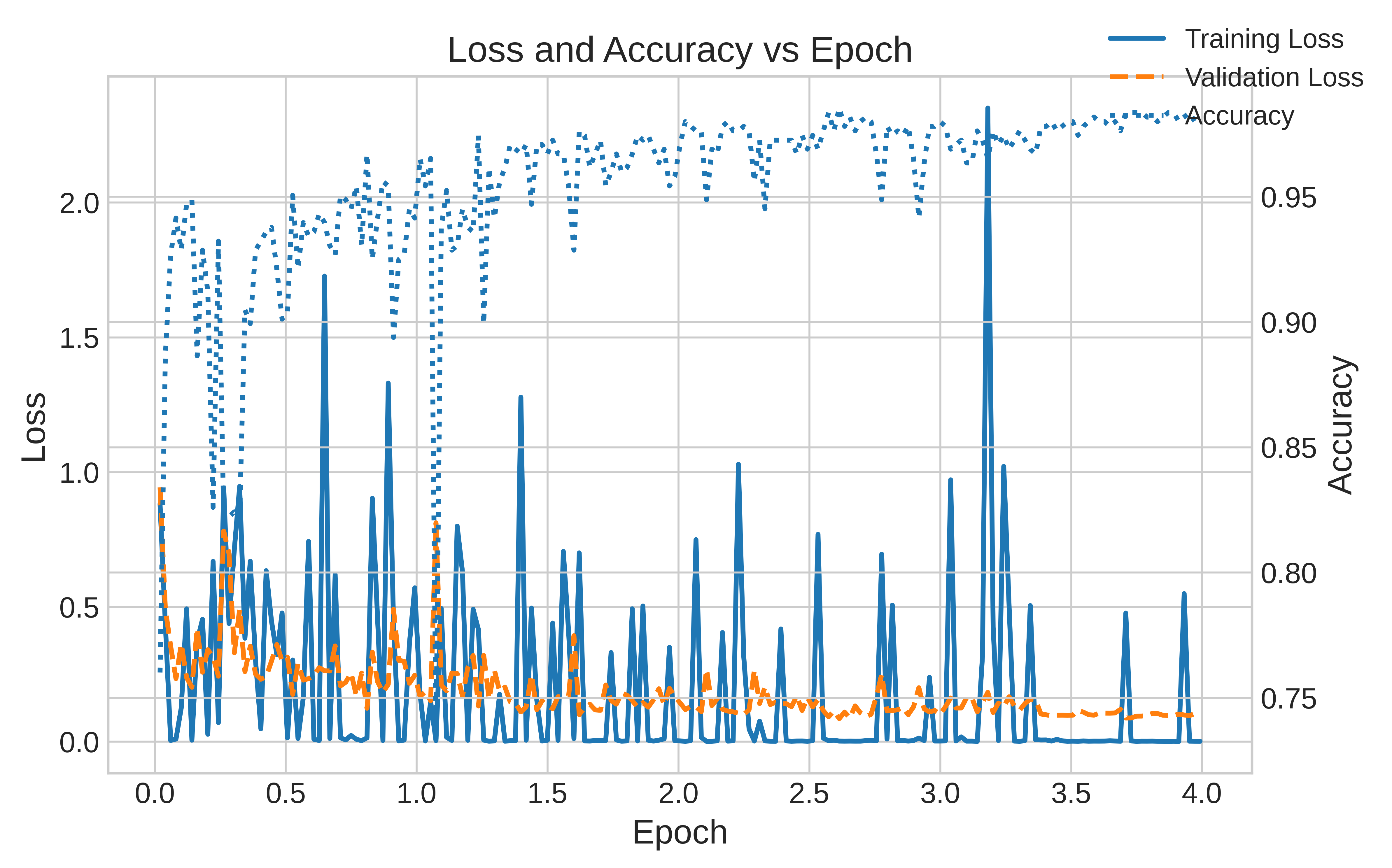}
  \caption{Joint loss/accuracy dynamics.}
\end{subfigure}

\vspace{0.5em}
\begin{subfigure}[b]{0.32\textwidth}
  \centering
  \includegraphics[width=\linewidth]{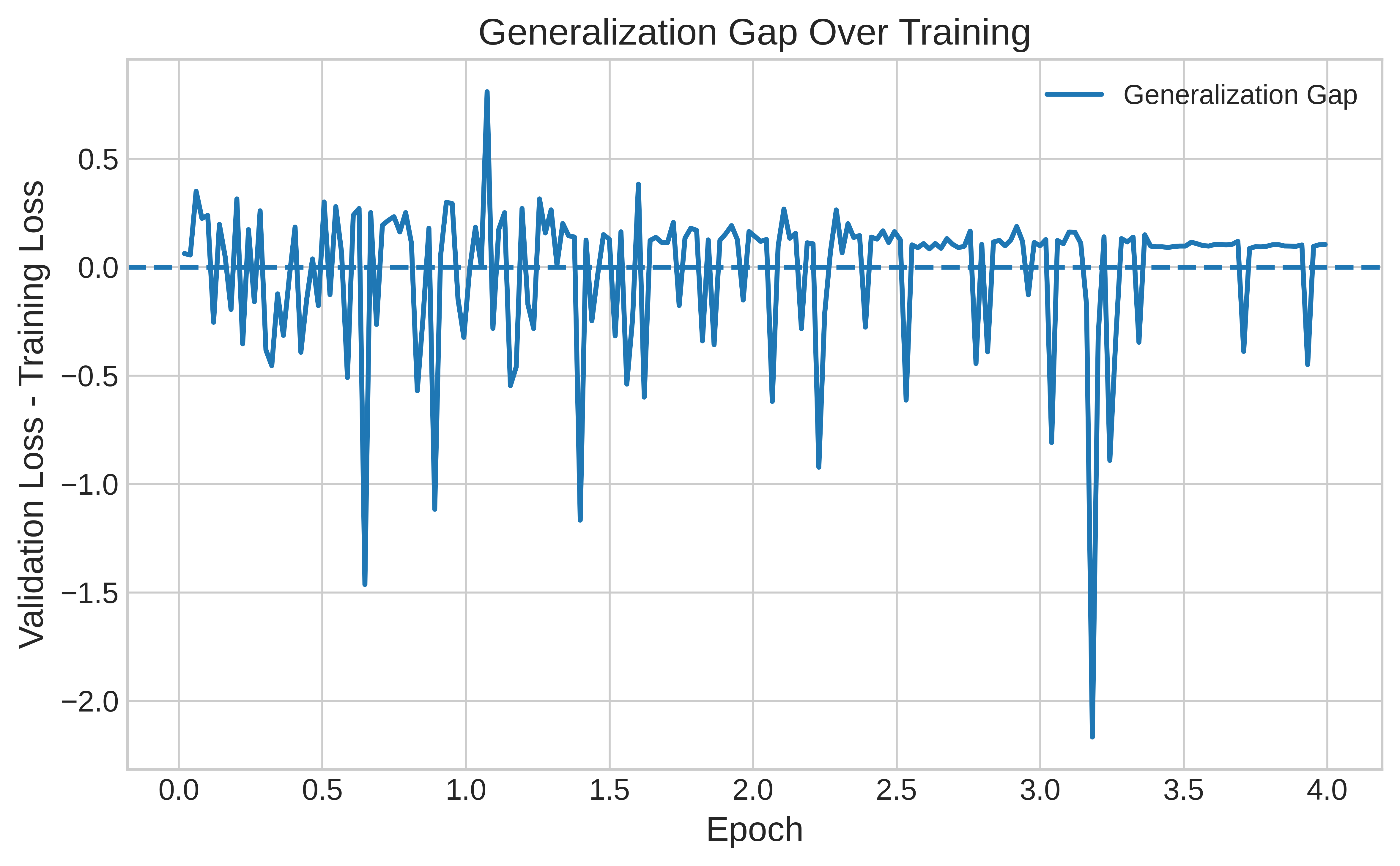}
  \caption{Generalization gap.}
\end{subfigure}
\begin{subfigure}[b]{0.32\textwidth}
  \centering
  \includegraphics[width=\linewidth]{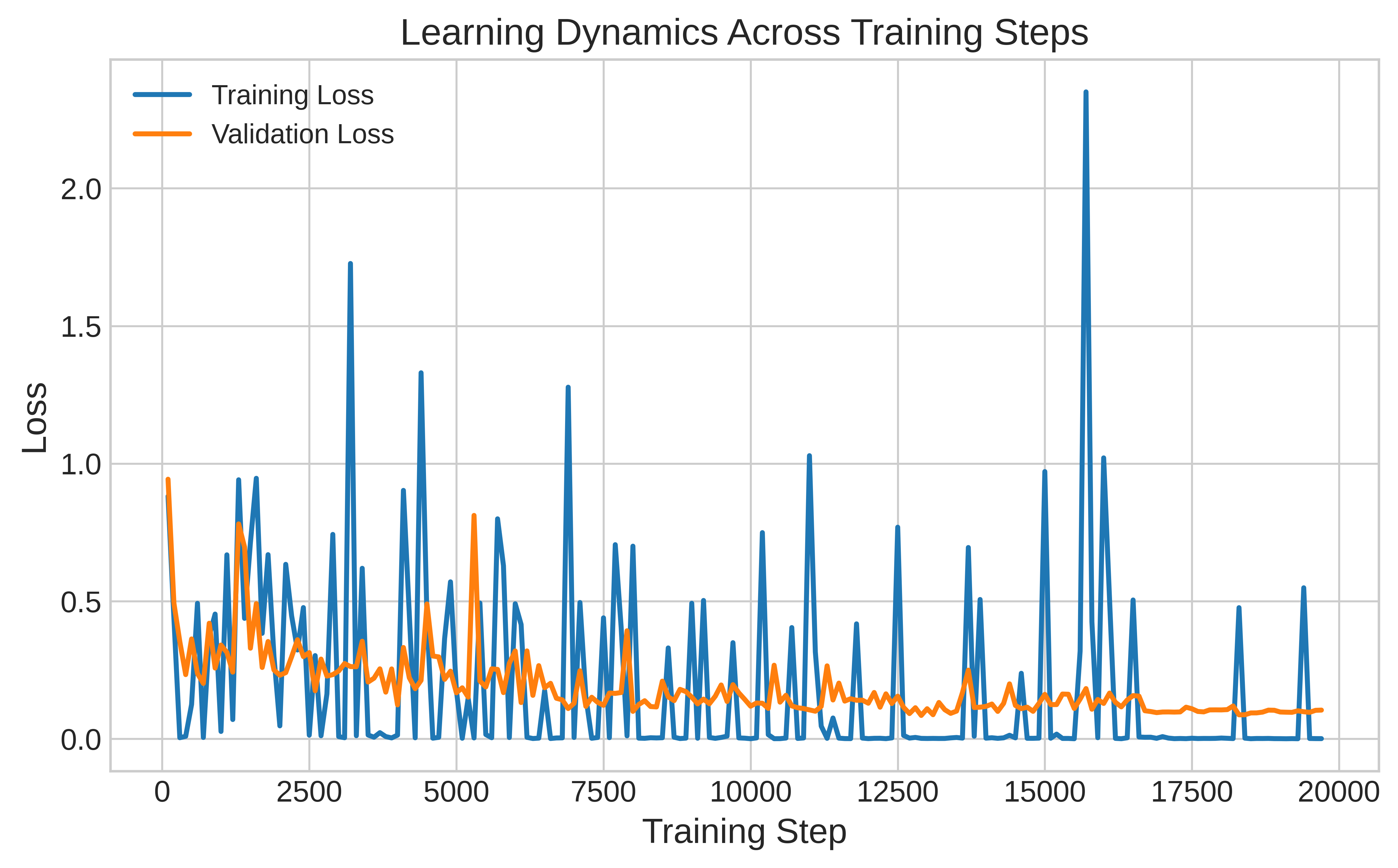}
  \caption{Learning dynamics by step.}
\end{subfigure}
\begin{subfigure}[b]{0.32\textwidth}
  \centering
  \includegraphics[width=\linewidth]{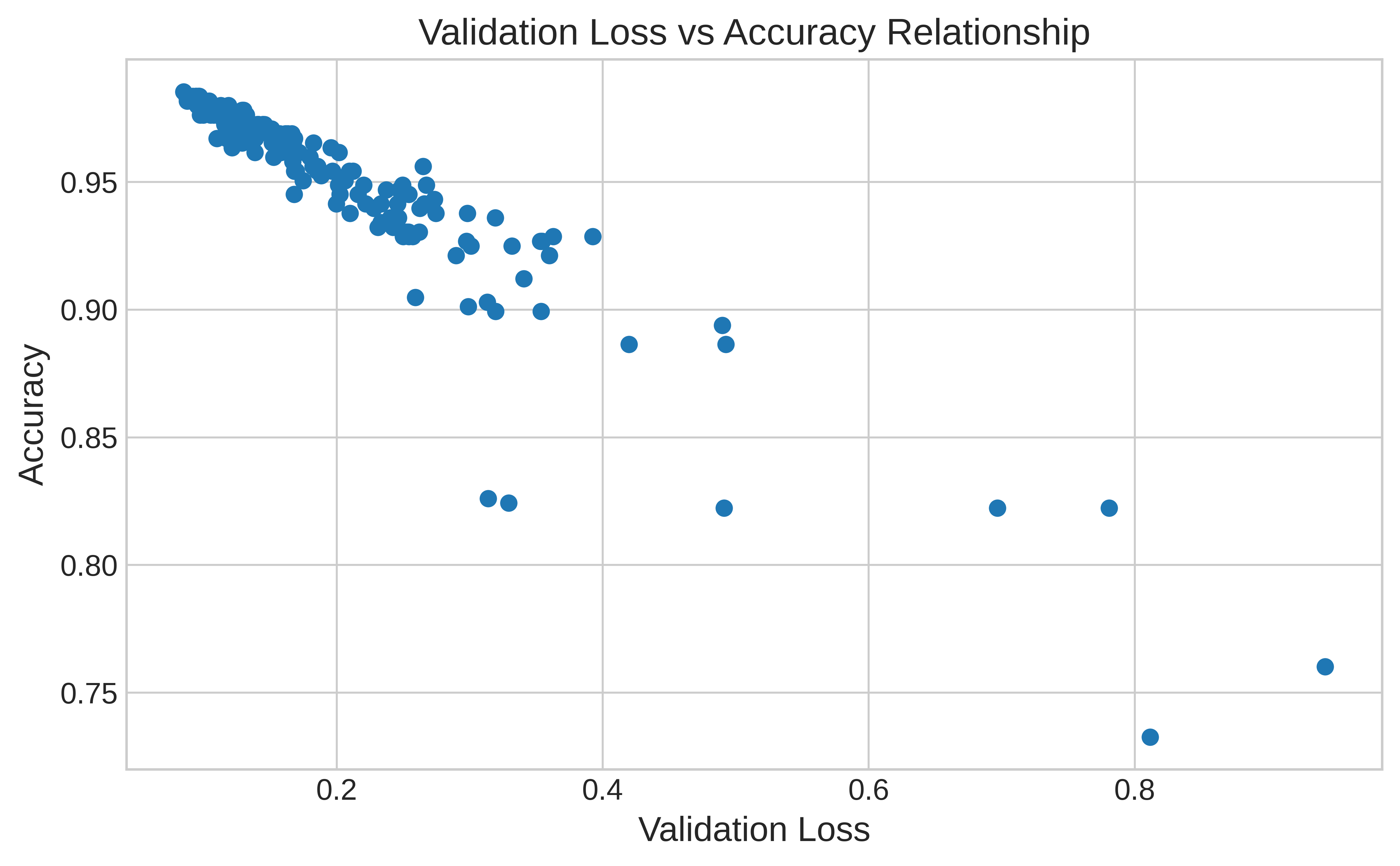}
  \caption{Validation loss vs.\ accuracy.}
\end{subfigure}
\caption{Training and validation behaviour of the ViT image classifier. Panels (a)--(c) show that loss decreases monotonically while accuracy rises to 98.53\%; (d) tracks the gap between training and validation performance, which remains small throughout; (e)--(f) highlight that convergence is fast (4 epochs) and that the final operating point lies in the regime of low validation loss and high accuracy.}
\label{fig:vit-training-overview}
\end{figure*}

Figure~\ref{fig:vit-training-overview} demonstrates that the classifier is slightly undertrained rather than overfit: both training and validation curves are still improving when training stops, and the generalization gap remains narrow. This behaviour is desirable for NovaLAD's ``useful vs.\ useless'' gate, because it prioritizes robust separation of truly informative figures from decorative imagery without memorizing the training set.

\subsection{DP-Bench Results}

Table~\ref{tab:results} shows the DP-Bench results for NovaLAD and other comparison systems (such the benchmark leaderboard and our runs). NovaLAD has the best TEDS (96.49\%) and NID (98.51\%) scores of all the other systems. Upstage has the second-highest scores (93.48\% TEDS, 97.02\% NID), AWS has the third-best scores (88.05\%, 96.71\%), Microsoft has the fourth-best scores (87.19\%, 87.69\%), Llamaparse has the fifth-best scores (74.57\%, 92.82\%), Unstructured has the sixth-best scores (65.56\%, 91.18\%), and Google has the seventh-best scores (66.13\%, 90.86\%). The improvements in TEDS show that table structure and content extraction are getting better, and the improvements in NID show that element identification and reading-order serialization are getting better. The average time per document for NovaLAD is 8.50 seconds, which is a little longer than the quickest APIs but without a GPU and with complete control over models and outputs. When needed, the pipeline can be modified (for example, by turning off insights or picture classification) to lower latency even more.

\begin{table}[H]
\centering
\caption{DP-Bench results: NovaLAD vs. other parsers. TEDS and NID $\uparrow$ (higher is better); Avg Time $\downarrow$ (lower is better).}
\label{tab:results}
\small
\begin{tabular}{lccc}
\toprule
System & TEDS $\uparrow$ & NID $\uparrow$ & Avg Time (s) $\downarrow$ \\
\midrule
NovaLAD $\star$ & \textbf{96.49} & \textbf{98.51} & 8.50 \\
Upstage & 93.48 & 97.02 & \textbf{3.79} \\
AWS & 88.05 & 96.71 & 14.47 \\
Microsoft & 87.19 & 87.69 & 4.44 \\
Llamaparse & 74.57 & 92.82 & 4.14 \\
Unstructured & 65.56 & 91.18 & 13.14 \\
Google & 66.13 & 90.86 & 5.85 \\
\bottomrule
\end{tabular}
\end{table}

\subsection{Novalad Results}

This section qualitatively demonstrates the behavior of NovaLAD on randomly selected document pages. The left subfigure shows the "input page image," and the right subfigure shows the "NovaLAD output" after the full pipeline has been run. This includes YOLO-based element and layout detection, ViT-based image usefulness classification, grouping and clustering of elements into layout-aware structures, and the final filtered bounding boxes rendered on top of the page. These pictures show how NovaLAD differentiates useful numbers from useless ones, keeps the structure of text and tables, and makes a clean, layout-aware copy of the original document.

\begin{figure}[H]
     \centering
     \begin{subfigure}[b]{0.48\textwidth}
         \centering
         \includegraphics[width=200pt]{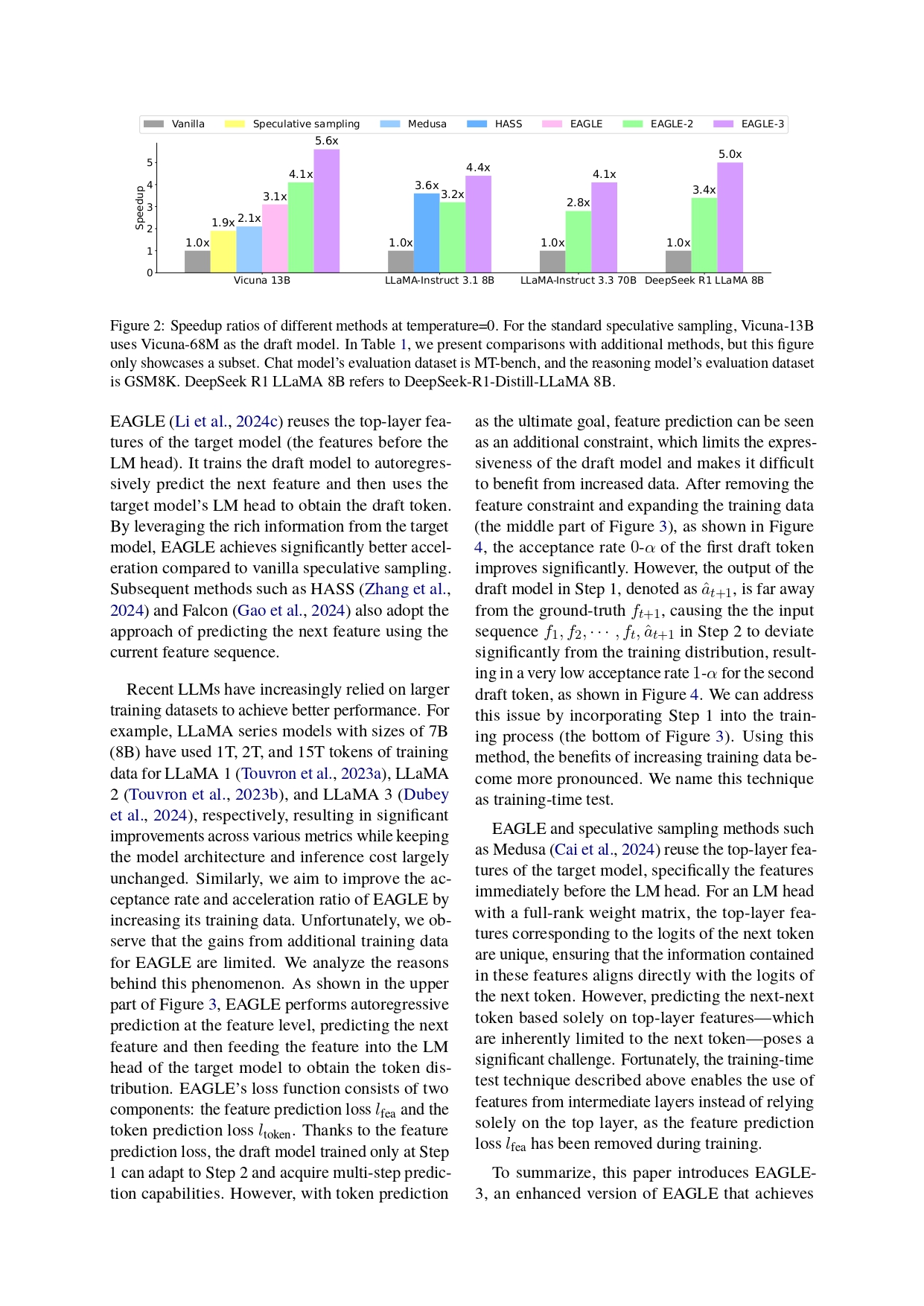}
         \caption{Input Image}
         \label{fig:loss_acc_1}
     \end{subfigure}
     \hfill 
     \begin{subfigure}[b]{0.48\textwidth}
         \centering
         \includegraphics[width=200pt]{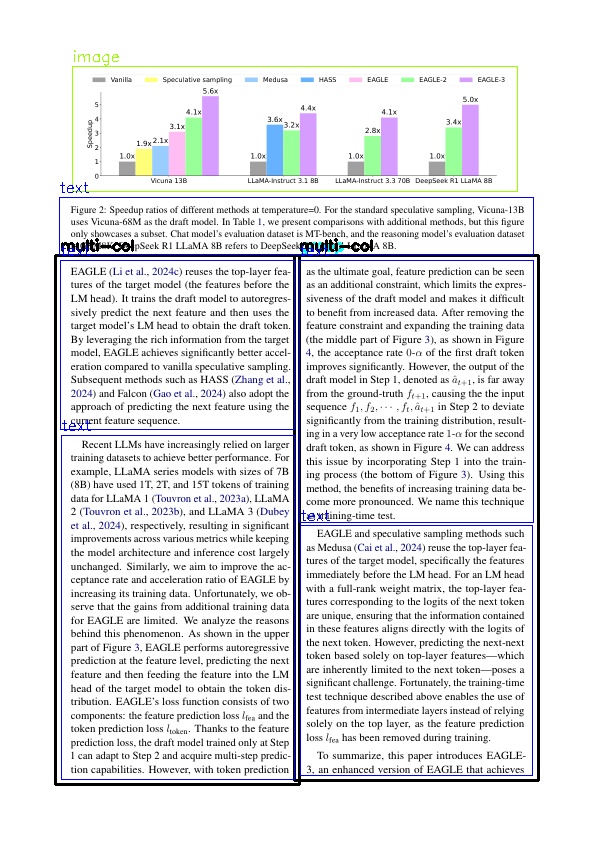}
         \caption{Novalad Result}
         \label{fig:loss_acc_2}
     \end{subfigure}
     \label{fig:novalad_output}
\end{figure}

\begin{figure}[H]
     \centering
     \begin{subfigure}[b]{0.48\textwidth}
         \centering
         \includegraphics[width=200pt]{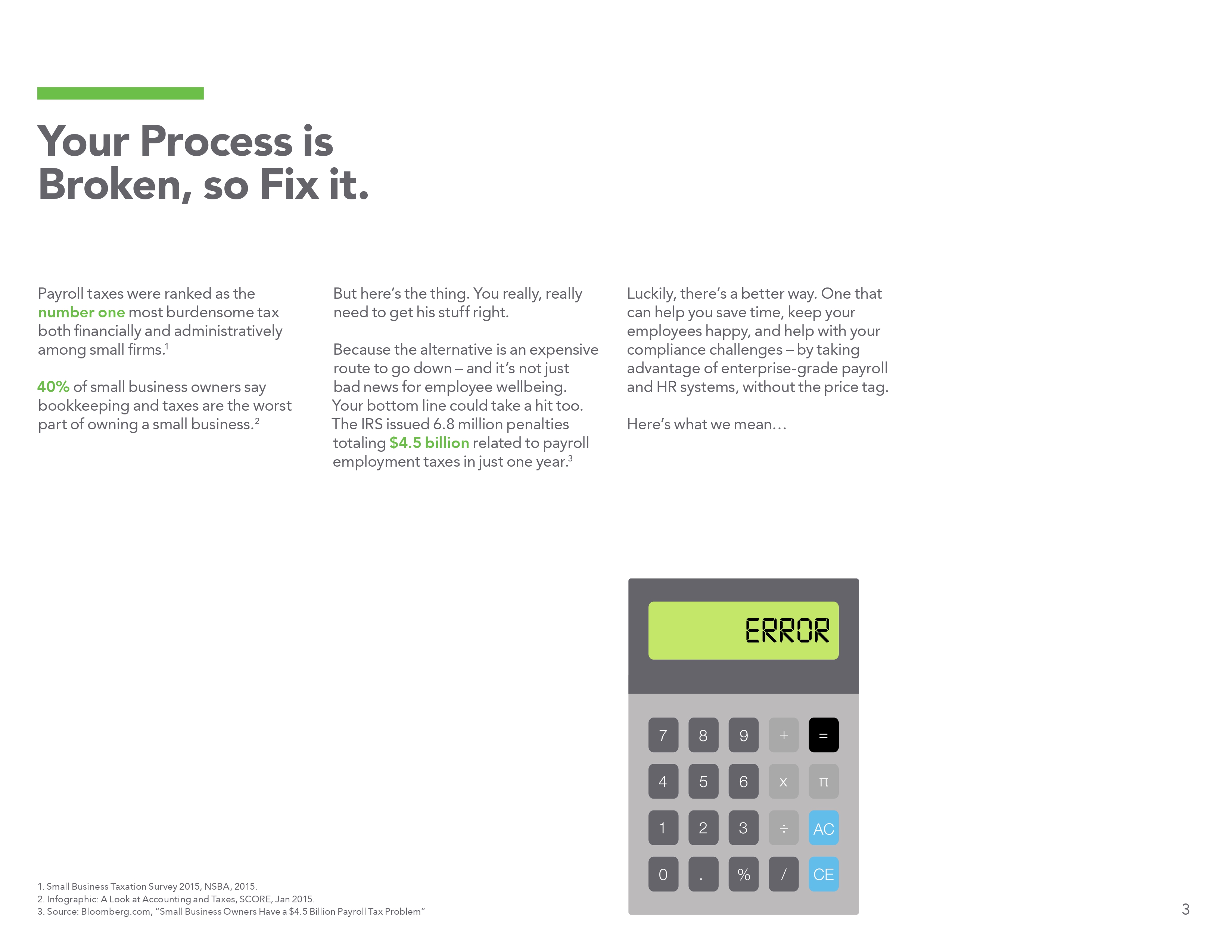}
         \caption{Input Image}
         \label{fig:loss_acc_1}
     \end{subfigure}
     \hfill 
     \begin{subfigure}[b]{0.48\textwidth}
         \centering
         \includegraphics[width=200pt]{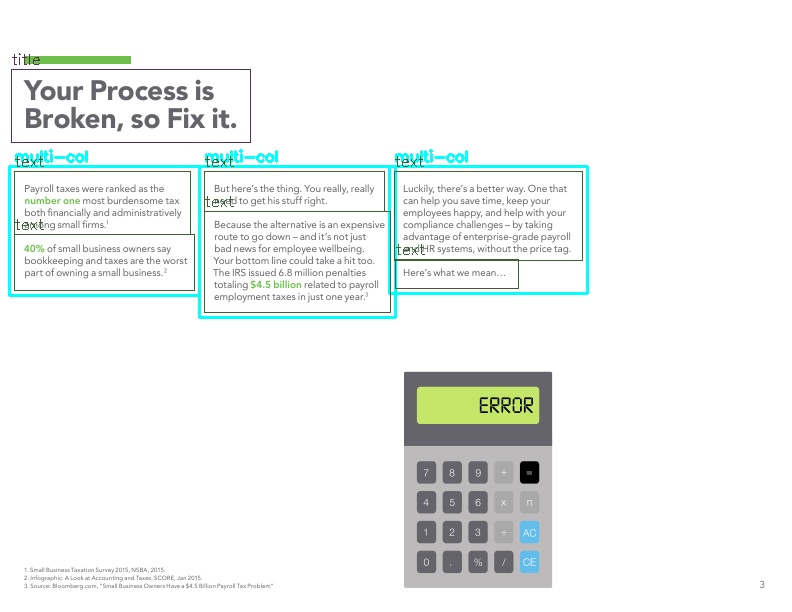}
         \caption{Novalad Result}
         \label{fig:loss_acc_2}
     \end{subfigure}
     \label{fig:overall_performance}
\end{figure}

\begin{figure}[H]
     \centering
     \begin{subfigure}[b]{0.48\textwidth}
         \centering
         \includegraphics[width=200pt]{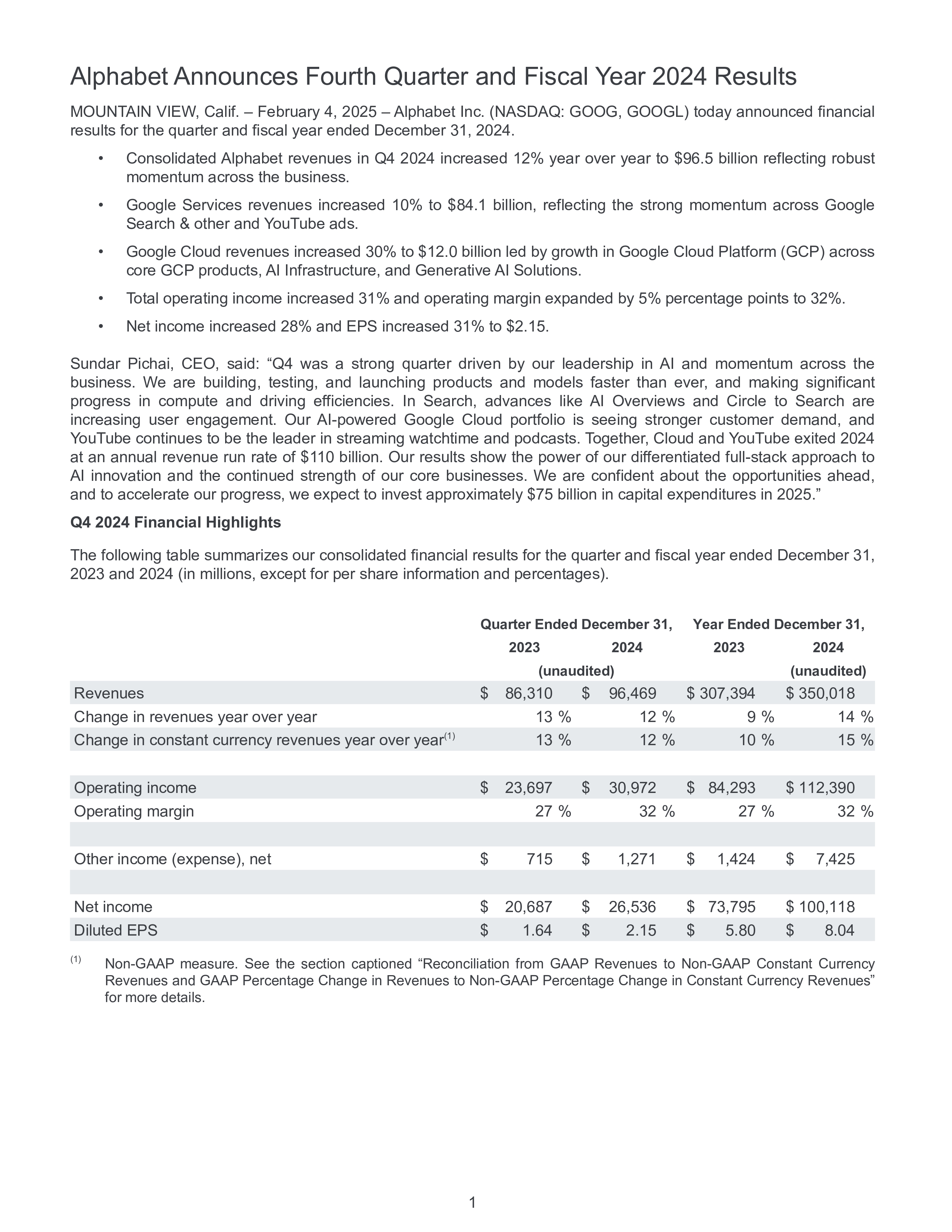}
         \caption{Input Image}
         \label{fig:loss_acc_1}
     \end{subfigure}
     \hfill 
     \begin{subfigure}[b]{0.48\textwidth}
         \centering
         \includegraphics[width=200pt]{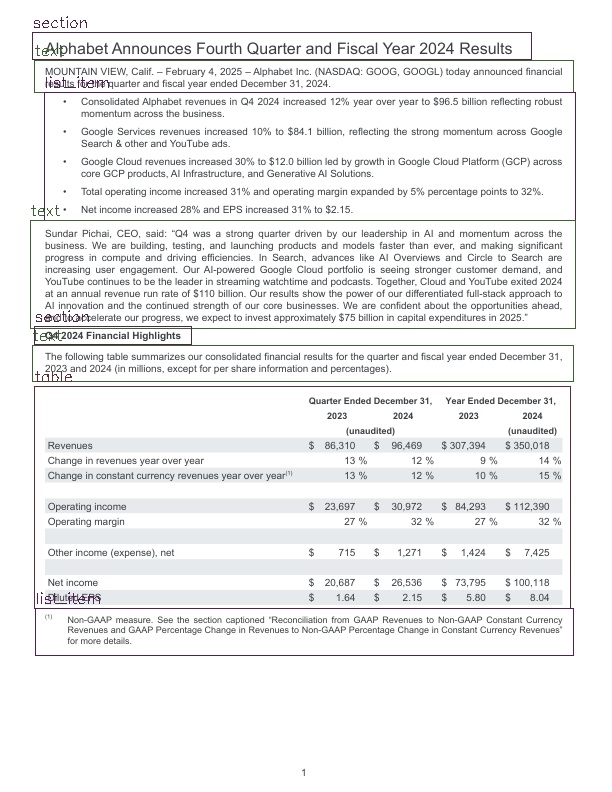}
         \caption{Novalad Result}
         \label{fig:loss_acc_2}
     \end{subfigure}
     \label{fig:overall_performance}
\end{figure}

\begin{figure}[H]
     \centering
     \begin{subfigure}[b]{0.48\textwidth}
         \centering
         \includegraphics[width=200pt]{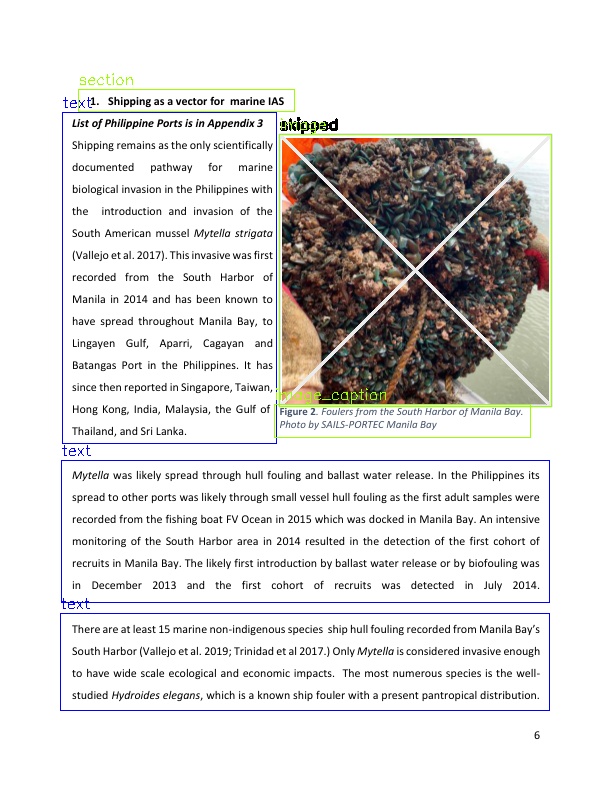}
         \caption{Input Image}
         \label{fig:loss_acc_1}
     \end{subfigure}
     \hfill 
     \begin{subfigure}[b]{0.48\textwidth}
         \centering
         \includegraphics[width=200pt]{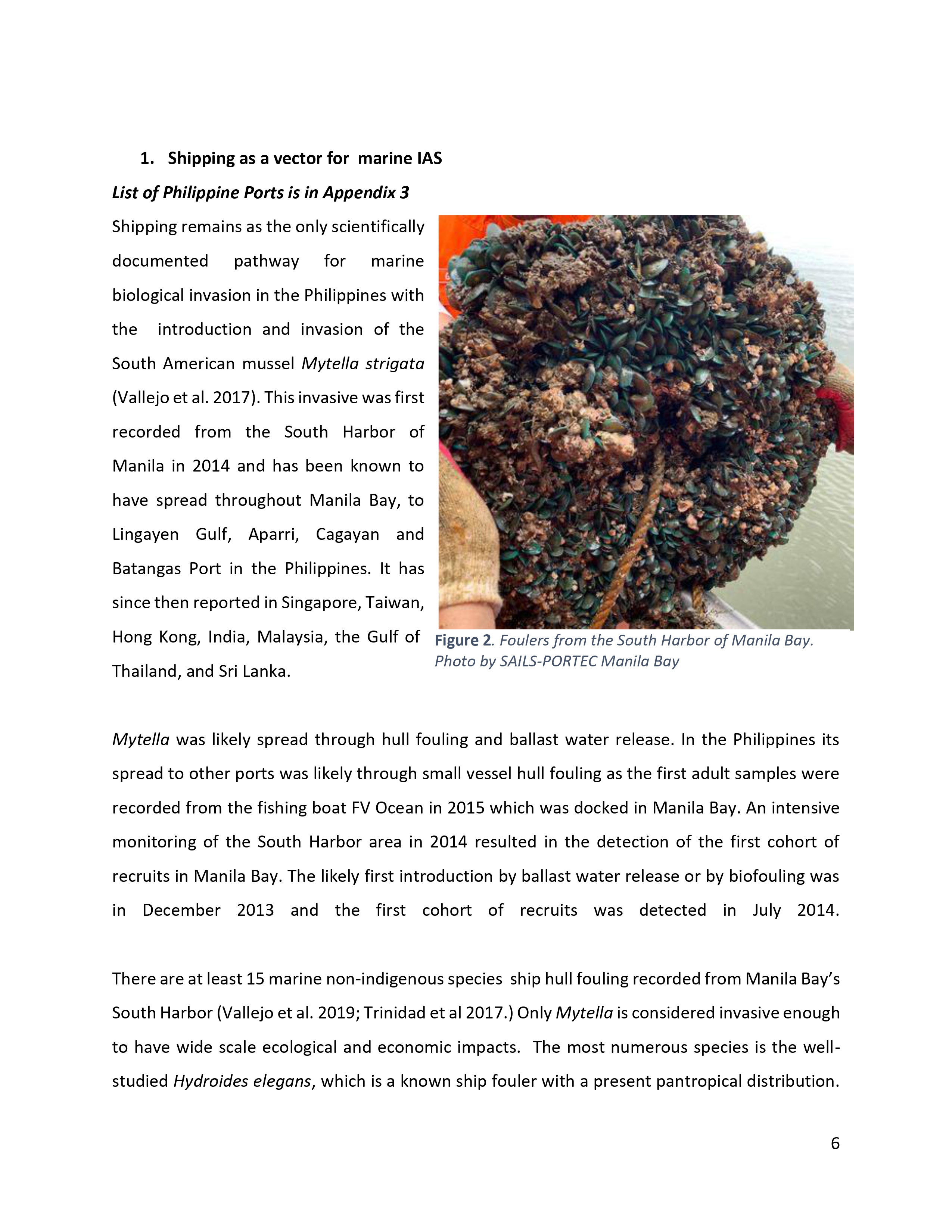}
         \caption{Novalad Result}
         \label{fig:loss_acc_2}
     \end{subfigure}
     \caption{Novalad Output post Element and Layout Detection}
     \label{fig:overall_performance}
\end{figure}

\section{Optimization and CPU-Based Design}
\label{sec:optimization}

\subsection{Speed and Parallelism}

NovaLAD is designed for \textbf{speed} through concurrency and minimal redundant work:

\begin{enumerate}
  \item \textbf{Per-page parallel detection}: Element model, layout model, and text-based category run in a single \texttt{ThreadPoolExecutor}; no sequential bottleneck.
  \item \textbf{Per-page parallel classification}: All image elements on a page are classified (ViT) in parallel; only those labeled \emph{useful} (plus all tables) are sent to the Vision LLM. This classifier gate reduces LLM calls and cost while keeping high-value figures and tables enriched.
  \item \textbf{Post-extraction parallel conversion}: Markdown, documents, and knowledge graph are produced from the same JSON in parallel, so the user gets all outputs with latency dominated by the slowest of the three, not the sum.
  \item \textbf{Single render per page}: One 300 DPI render; all detectors share the same image. Coordinate scaling (image $\leftrightarrow$ PDF) is done once per box.
  \item \textbf{Lightweight grouping}: DBSCAN and sort-based grouping are CPU-only and fast; no extra neural forward passes for grouping.
\end{enumerate}

\subsection{CPU-Based Execution}

You don't need a GPU for the pipeline. YOLOv10, EasyOCR, ViT, and the text classifier can all run on a CPU (PyTorch \texttt{device = "cuda" if available else "cpu"}); settings can force \texttt{TORCH\_DEVICE = "cpu"} for deployment on servers that only have CPUs. This lowers costs and makes operations easier, thus NovaLAD can be used in air-gapped or resource-limited settings while still getting the best DP-Bench scores. Optional steps like converting LibreOffice files or getting vision-language insights may use external services or subprocesses, but they are not necessary for core extraction. The detection, grouping, and conversion approach is self-contained and CPU-bound.

\subsection{Implementation Notes}

The codebase is modular: \texttt{extractors/pdf} (load, metadata), \texttt{convertors/pdf} (extract\_pdf: detect, then \emph{image classifier gate} for image/figure elements, then Vision LLM only for useful images and tables, then group, order, header/footer), \texttt{models} (YOLOv10, EasyOCR, ViT for usefulness classification, transformer for document category, with \texttt{run\_detection} and \texttt{run\_classification}), \texttt{convertors/markdown}, \texttt{convertors/documents}, \texttt{convertors/graph} (parallel conversions), and \texttt{job.py} (orchestration). Some of the dependencies are pypdfium2, PyMuPDF, ultralytics (YOLOv10), supervision, EasyOCR, OpenCV, transformers, torch, scikit-learn (DBSCAN, scalers), thefuzz, and OpenAI (for insights).

\section{Conclusion}
\label{sec:conclusion}

We contended that document extraction is an essential element of AI and data intelligence, particularly for RAG and generative AI, as it determines the quality of the input for all subsequent models. We showed off NovaLAD, a complete document parsing pipeline that uses YOLOv10-based element and layout detection, EasyOCR for tables and figures, a ViT image classifier that only sends useful image and figure elements to the Vision LLM for title, summary, and structured content, and rule-based grouping and reading-order serialization. NovaLAD can make papers, knowledge graphs, and JSON, Markdown, and RAG-ready files that work with DP-Bench at the same time. On the DP-Bench benchmark (upstage/dp-bench), NovaLAD gets top scores of \textbf{TEDS 96.49\%} and \textbf{NID 98.51\%}, showing that a single, well-optimized pipeline can do better than both commercial and open-source options. NovaLAD is quick (it runs in parallel at every level), CPU-friendly, and cost-aware (the image classifier gate only lets Vision LLM calls get to useful figures and tables). Future research may investigate more extensive layout models, supplementary languages for OCR, and enhanced interaction with RAG retrieval measures.

To see the code for novalad : \url{https://github.com/novaladai/novalad}

\balance

\begin{thebibliography}{99}
\bibitem{dpbench}
DP-Bench: Document Parsing Benchmark. Hugging Face Dataset: upstage/dp-bench. \url{https://huggingface.co/datasets/upstage/dp-bench}.

\bibitem{rag}
Lewis, P. et al. Retrieval-Augmented Generation for Knowledge-Intensive NLP. NeurIPS 2020.

\bibitem{constraint_layout}
M. Breuel. Two geometric algorithms for layout analysis. DAS 2002.

\bibitem{pdfminer}
Y. Tanaka. PDFMiner: PDF parser and analyzer. \url{https://github.com/pdfminer/pdfminer}.

\bibitem{reading_order_survey}
A. Antonacopoulos et al. A survey of document image analysis and recognition. ICDAR 2009; reading order and layout analysis.

\bibitem{layoutlm}
Y. Xu et al. LayoutLM: Pre-training of Text and Layout for Document Image Understanding. KDD 2020.

\bibitem{docbank}
M. Li et al. DocBank: A Benchmark Dataset for Document Layout Analysis. COLING 2020.

\bibitem{yolo_doc}
Ultralytics YOLO. \url{https://github.com/ultralytics/ultralytics}. Document detection adaptations.

\bibitem{detr_doc}
N. Carion et al. End-to-End Object Detection with Transformers. ECCV 2020.

\bibitem{teds}
Z. Zhong et al. Image-based table recognition: data, model, and evaluation. ECCV 2020 (TEDS metric).

\bibitem{table_transformer}
B. Smock et al. PubTables-1M: Towards comprehensive table extraction from unstructured documents. CVPR 2022.

\bibitem{tablestr}
Table structure recognition. Various GNN/transformer approaches; see DP-Bench and TEDS references.

\bibitem{textract}
Amazon Textract. \url{https://aws.amazon.com/textract/}.

\bibitem{docai}
Google Cloud Document AI. \url{https://cloud.google.com/document-ai}.

\bibitem{ms_read}
Microsoft Document Intelligence (formerly Form Recognizer). \url{https://azure.microsoft.com/en-us/products/ai-services/ai-document-intelligence}.

\bibitem{unstructured}
Unstructured.io. \url{https://unstructured.io/}. Open-source document parsing and chunking.
\end{thebibliography}
\end{document}